\documentclass{article} 

\usepackage{arxiv_preamble}
\usepackage{hyperref} 
\usepackage[most]{tcolorbox}
\usepackage{xcolor}

\definecolor{mdbg}{HTML}{F2F4F7}   
\definecolor{mdframe}{HTML}{E5E7EB}
\definecolor{mdblue}{HTML}{4F46E5} 

\definecolor{YaleBlue}{HTML}{00356B} 
\definecolor{PaperRed}{HTML}{A51C30} 

\definecolor{MorandiGreen}{HTML}{A2A896} 
\definecolor{MorandiOrange}{HTML}{D5A37D} 
\definecolor{MorandiRed}{HTML}{C48273}   
\definecolor{MorandiBlue}{HTML}{8DA4B3}  
\definecolor{MorandiGray}{HTML}{9E9E9E}   
\definecolor{MorandiPurple}{HTML}{B3A6B3} 
\definecolor{PaperGreen}{HTML}{006400} 

\newtcbox{\mdcode}{on line, enhanced,
  colback=mdbg, colframe=mdframe, coltext=mdblue,
  boxrule=0.4pt, boxsep=0.2ex, left=0.6ex, right=0.6ex,
  top=0.2ex, bottom=0.15ex, arc=3pt,
  fontupper=\ttfamily, tcbox raise base}
\usepackage{letltxmacro}

\newtcbox{\mdcodev}{on line, enhanced, verbatim,
  colback=mdbg, colframe=mdframe, coltext=mdblue,
  boxrule=0.4pt, boxsep=0.2ex, left=0.6ex, right=0.6ex,
  top=0.2ex, bottom=0.15ex, arc=3pt, tcbox raise base}
\usepackage{ifthen}
\newcommand{\compileversion}{bodyappendix} 

\title{\huge MultiView-Bench: A Diagnostic Benchmark for World-Centric Multi-View Integration in VLMs}

\author{%
  Hantao Zhang$^{1,\dagger}$ \\
  \texttt{hantao.zhang@yale.edu} \\
  \and
  Jinru Sui$^{2}$ \\
  \texttt{s2680116@sms.ed.ac.uk} \\
  \and
  Ed Li$^{1}$ \\
  \texttt{ed.li@yale.edu} \\
  \and
  Dirk Bergemann$^{1}$ \\
  \texttt{dirk.bergemann@yale.edu} \\
  \and
  Zhuoran Yang$^{1}$ \\
  \texttt{zhuoran.yang@yale.edu} \\
}

\date{} 

\hypersetup{
  pdftitle={MultiView-Bench: A Diagnostic Benchmark for World-Centric Multi-View Integration in VLMs},
  pdfauthor={Hantao Zhang, Jinru Sui, Ed Li, Dirk Bergemann, Zhuoran Yang}
}

\begin{document}

\renewcommand{\thefootnote}{}
\footnotetext{$^\dagger$ Correspondence: \texttt{hantao.zhang@yale.edu}}
\renewcommand{\thefootnote}{\arabic{footnote}}

\ifthenelse{\equal{\compileversion}{bodyonly}}{
    \maketitle

    \section*{Introduction}
    \label{sec:intro}

    Recent advances in Large Language Models (LLMs) \citep{brown2020language, achiam2023gpt, hurst2024gpt, touvron2023llama} and Vision-Language Models (VLMs) \citep{radford2021learning, li2022blip, team2023gemini, dai2023instructblip} have demonstrated remarkable progress in complex perceptual and reasoning tasks, including spatial navigation \citep{zhou2024navgpt, yamada2023evaluating} and image understanding \citep{dosovitskiy2020image}. Their strong generalization capabilities, coupled with emergent reasoning skills, make them compelling candidates for \textit{cognitive systems} that integrate perception and strategic planning \citep{bubeck2023sparks}. When equipped with appropriate tools and scaffolding, such systems have shown promise in robotics control \citep{zitkovich2023rt}, 3D modeling \citep{hu2024scenecraft, gu2025blendergym}, and image editing \citep{huang2024blenderalchemy}.

However, effectively solving many of these tasks fundamentally depends on the ability to \textit{perceive and reason about scenes from multiple viewpoints} \citep{edelman1998representation, bulthoff1992psychophysical}.  Humans naturally perform multi-angle observations to construct coherent mental models of objects, resolving perceptual ambiguities that arise from single viewpoints \citep{shepard1971mental}. This ability is crucial when assembling complex objects, where each component must be rotated and inspected from multiple viewpoints to determine how it connects with others. In contrast, a single static image often fails to convey critical structural or relational details necessary for accurate reasoning and manipulation, underscoring the importance of \textit{multi-view perception} in spatial cognition \citep{marr2010vision}.

A possible workaround involves geometric representations such as point clouds, meshes, or voxels \citep{qi2017pointnet, wu20153d, mescheder2019occupancy}, which encode precise 3D coordinates and shapes. However, processing such low-level geometric data typically requires specialized encoders \citep{qi2017pointnet++, wang2019dynamic} and lacks the broad generalization of LLM/VLM-based approaches. Moreover, these representations diverge from the modality of human visual input, limiting their interpretability for \textit{human-readable reasoning}.

Current multi-view spatial reasoning benchmarks \citep{wang2025site, daxberger2025mm, yin2025spatial, zhang2025flatland} primarily assess egocentric spatial reasoning such as perspective-taking, view-dependent navigation, or dynamic viewpoint transformation. These benchmarks evaluate a model's ability to map pixels from one 2D plane to another or to predict the visual consequences of camera motion without targeting view-invariant (world-centric) spatial understanding and holistic 3D grounding. Such world-centric reasoning ability is a prerequisite for mechanical engineering and operating 3D modeling software (e.g. Blender) where the agent needs to manipulate in global space, independent of where the user is looking.

In this work, we fill this gap by introducing \textbf{MultiView-Bench}, a multi-view spatial reasoning benchmark that explicitly requires the model to integrate multiple viewpoints to reconstruct a static, global mental model of the scene. We ground this by providing a visible, fixed global coordinate system, compelling the model to decouple object positioning from the camera's transient perspective. Our contributions are as follows:

\begin{itemize}
\item We introduce \textbf{MultiView-Bench}, a diagnostic benchmark for evaluating VLMs' ability to understand the 3D global coordinate system in 3D modeling software and integrate multi-view observations into a coherent world-centric 3D scene understanding. Our benchmark serves as a prerequisite test VLMs should pass before deployment to mechanical engineering and 3D modeling. The main benchmark contains 5 task variants each with 100 tasks where we conduct a systematic evaluation of state-of-the-art VLMs.
\item Alongside the benchmark, we provide a flexible \textbf{data generation pipeline} that allows researchers to easily extend the dataset with new 3D assets, task variants, and viewpoint configurations. We created 20 more different task variants each with 100 tasks and revealed key failure modes, biases, and limitations in VLMs' multi-view spatial reasoning capabilities through controlled experiments.
\item We propose \textbf{ViewNavigator}, a multi-agent framework that models perception, planning, and belief-updating. ViewNavigator consistently and significantly enhances the performance of underlying VLMs on MultiView-Bench, demonstrating its potential as a plug-and-play reasoning scaffold for future vision--language systems.
\end{itemize}

\section{Related Works}
\paragraph{VLM Benchmarks.}  
Numerous benchmarks have been developed to evaluate VLM capabilities. Foundational datasets such as ShapeNet \citep{chang2015shapenet} and ModelNet \citep{wu20153d} focus on 3D object recognition and reconstruction. CLEVR \citep{johnson2017clevr} targets compositional reasoning in synthetic images. More recent work has extended to spatial reasoning: SpatialRGPT \citep{cheng2024spatialrgpt} and OmniSpatial \citep{jia2025omnispatial} incorporate perspective-taking and dynamic reasoning, but remain primarily \textit{single-view} in nature. 

While other multi-view benchmarks exist, they often prioritize different cognitive tasks. InternSpatial \citep{deng2025internspatial} is restricted to rotation estimation. ViewSpatial-Bench \citep{li2025viewspatial}, SITE \citep{wang2025site}, SPAR \citep{zhang2025flatland} and MindCube \citep{yin2025spatial} focus on egocentric--allocentric viewpoints transformations and navigation dynamics that requires sequential feature matching. Crucially, these benchmarks do not require the model to integrate multiple concurrent views into a single, coherent global representation. Finally, benchmarks like IR3DBench \citep{liu2025ir3d} and MM-Spatial \citep{daxberger2025mm} test precise 3D coordinate estimation. However, IR3DBench is restricted to single views which allows multiple plausible configurations to produce the same image and MM--Spatial \citep{daxberger2025mm} relies on explicit depth maps, which hinders scalability compared to pure RGB image reasoning. We summarize the distinctions from these image-only 3D multi-view VLM benchmarks in Table~\ref{tab:benchmark_comparison}. 

\paragraph{3D Spatial Reasoning with LLM/VLM Agents.}  
Agentic systems leveraging LLMs and VLMs have recently demonstrated impressive performance on 3D tasks, including open-world gaming \citep{wang2023voyager, yao2023react}, procedural scene generation \citep{hu2024scenecraft, huang2024blenderalchemy}, and LEGO assembly \citep{yamada2025l3go, pun2025generating}. In such systems, LLMs often act as \textit{planners}, while VLMs serve as perceptual modules that transform raw visual inputs into structured descriptions. These pipelines work well when single-image perception suffices (e.g., block-based abstractions in Minecraft), but break down when tasks require precise geometric reasoning over multiple views \citep{chen2024spatialvlm, hong20233d}.

While some works attempt to enhance VLM 3D reasoning \citep{cheng2024spatialrgpt, chen2024spatialvlm, hong20233d}, they typically focus on VQA-style setups without extending to applications that require integrated 3D perception and planning. MultiView-Bench is designed precisely to call for the awareness of VLMs' limitations in multi-view integration and to serve as a selection criterion when building VLM-based agents for operating 3D software in 3D asset generation and mechanical engineering.
\begin{table}[h!]
\centering
\caption{Distinctions between existing image-only multi-view VLM benchmarks \citep{wang2025site, yin2025spatial, zhang2025flatland} and MultiView-Bench.}
\label{tab:benchmark_comparison}
\renewcommand{\arraystretch}{1.2} 
{
\begin{tabular}{p{0.15\linewidth} p{0.38\linewidth} p{0.38\linewidth}}
\toprule
\textbf{Feature} & \textbf{Existing Multi-View Benchmarks} & \textbf{MultiView-Bench} \\
\midrule
\textbf{Environment} & \textbf{Photorealistic / Real-World Scenes} \newline High visual noise, complex textures, messy lighting. & \textbf{Synthetic / Blender-GUI} \newline Clean, schematic geometry with visible global axes/grids. \\
\midrule
\textbf{Reference Frame} & \textbf{View-dependent} \newline Relative to the camera (Left, Right, Forward). & \textbf{View-invariant} \newline Relative to a fixed global coordinate system. \\
\midrule
\textbf{Reasoning Type} & \textbf{Transformation} \newline $View_1 \rightarrow \text{Transformation} \rightarrow View_2$ & \textbf{Intersection} \newline $View_1 + View_2 + ... \rightarrow \text{Global State}$ \\
\midrule
\textbf{Primary Goal} & \textbf{End-to-End Performance} \newline Testing general robustness in noisy, real-world environments. & \textbf{Diagnostic Prerequisite} \newline "Unit testing" precise spatial logic before deployment. \\
\midrule
\textbf{Domain} & \textbf{Navigation \& Embodied AI} \newline "Where do I go next?" \newline "What will I see next?" & \textbf{Engineering \& Manipulation} \newline "Where is this part located exactly?" \newline "Where should I put the next block?"\\
\bottomrule
\end{tabular}
}
\end{table}

\section{Motivation: Furniture Assembly as a Case Study}


\begin{figure*}[h]
\centering
\begin{subfigure}{0.48\textwidth}
    \centering
    \includegraphics[width=\textwidth]{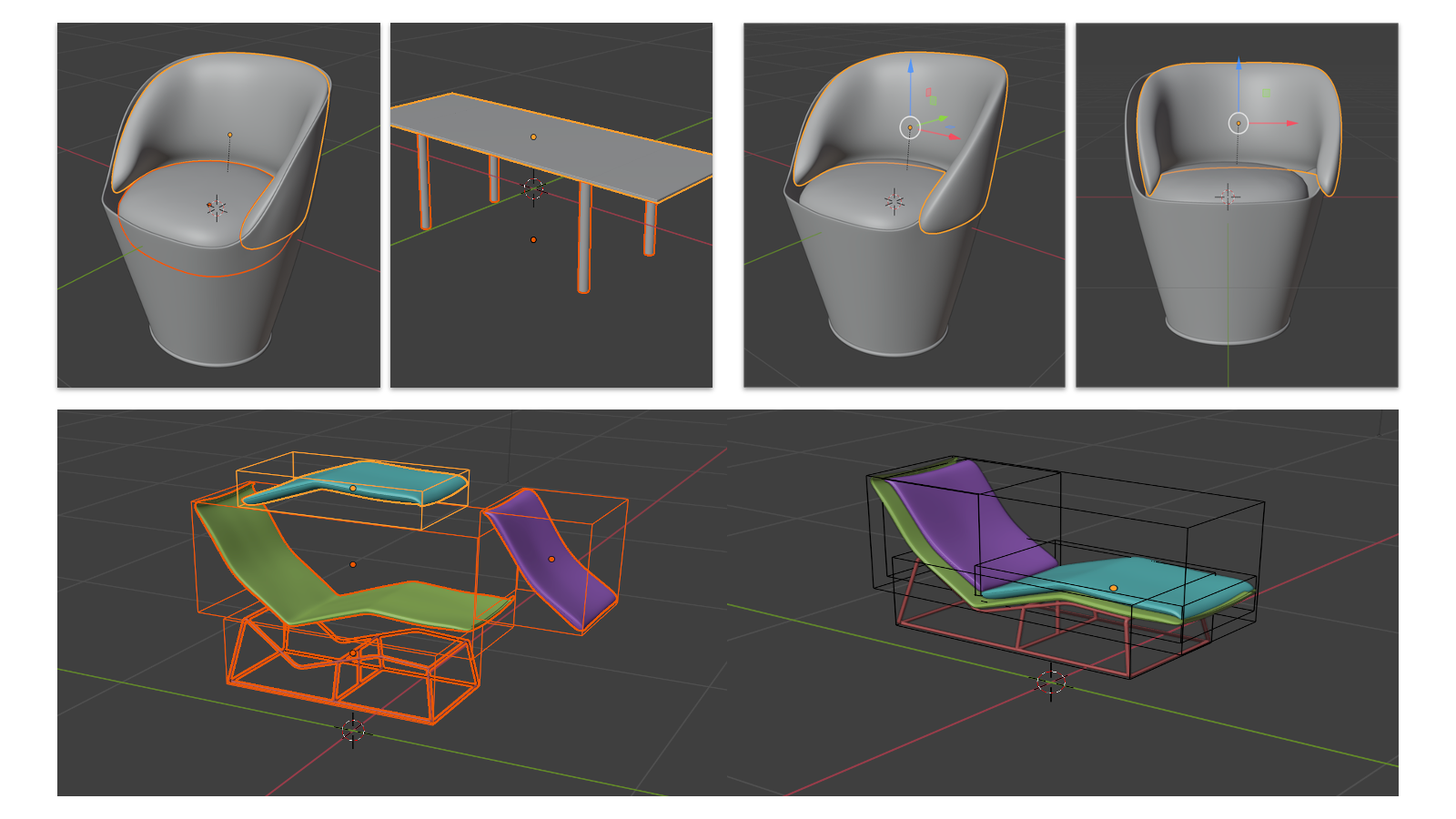}
    \caption{}
    \label{fig:furniture_example}
\end{subfigure}
\hfill
\begin{subfigure}{0.50\textwidth}
    \centering
    \includegraphics[width=\textwidth]{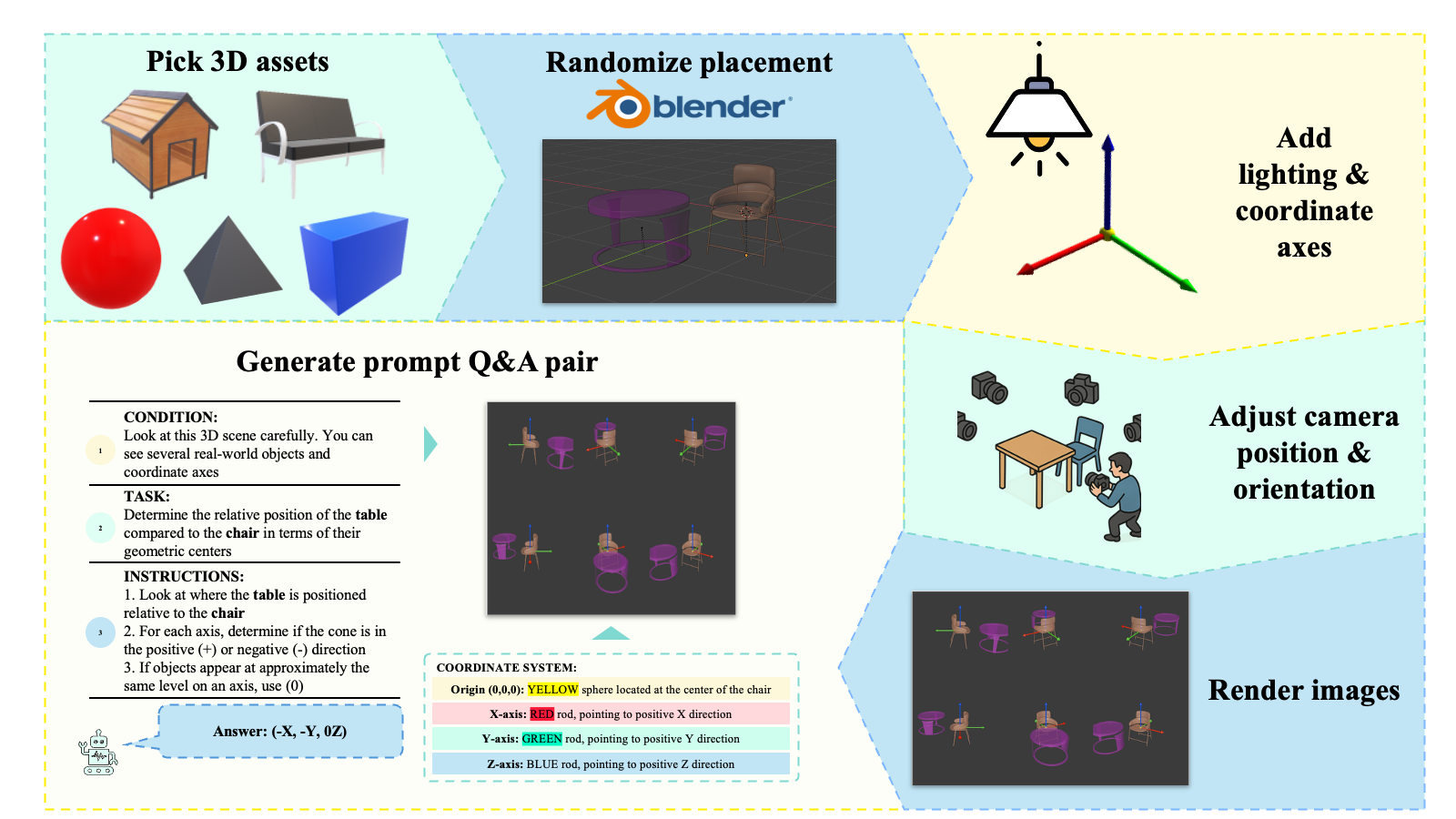}
    \caption{}
    \label{fig:data_generation_pipeline}
\end{subfigure}
\caption{(a) Furniture Assembly Example. \textbf{Top Left}: Many real-world objects do not lend themselves to simple natural language description. The table on the right can be described using fundamental convex shapes and their bounding boxes but the chair on the left has non-convex parts without an analytical expression. Thus it is preferable to include visual information. \textbf{Top Right}: Oftentimes, single-view observation leads to visual misconceptions and does not reveal certain alignment issues. The chair looks well-assembled in the view shown on the left but when it turns to the view shown on the right we see the backrest is slightly misplaced in the X-axis. \textbf{Bottom}: Only using the bounding box dimensions, we are unable to assemble furniture that has non-convex parts. In the left instance, the bounding boxes are perfectly aligned but the assembly is problematic. A good assembly example is shown on the right. (b) Data generation pipeline of MultiView-Bench with one example task for VLMs.}
\label{fig:combined_furniture_pipeline}
\end{figure*}

As LLMs and VLMs are increasingly deployed as agents to control software tools, measuring their proficiency within these interfaces is critical. To illustrate the necessity and practical importance of our benchmark, we motivate our study through a furniture part assembly task in Blender GUI (Figure~\ref{fig:furniture_example}). In this setting, a collection of labeled components (e.g., legs, table tops, backrests) must be connected and arranged to form a functional piece of furniture. Solving this task naturally demands multi-view perception, 3D spatial reasoning, and common-sense knowledge about how objects are typically used and combined. At a minimum, the following three core abilities are required:

\begin{enumerate}
\item \textbf{Part Identification and Semantic Reasoning:} Identifying distinct furniture components and inferring their functionality using combined visual and semantic reasoning.
\item \textbf{Assembly Planning:} Formulating a coherent and executable plan by determining the correct assembly sequence and how components interconnect.
\item \textbf{Assembly Execution:} Precisely placing each component and executing assembly actions.
\end{enumerate}

Recent LLM-based approaches have shown strong capabilities in environments such as Minecraft or Blender \citep{yamada2025l3go, pun2025generating, wang2023voyager}. However, these assemblies involve standardized, uniform blocks that lend themselves to lossless bounding-box (length, width, height) representations. In contrast, furniture assembly poses significantly greater complexity due to irregular, non-convex shapes that defy concise, lossless linguistic descriptions. Hence, purely semantic or bounding-box representations are inadequate for precise assembly tasks involving intricate realistic parts.

\begin{tcolorbox}[colback=MorandiBlue!10!white, colframe=MorandiBlue!80!black]
\textbf{Core Motivation:} Tasks like furniture assembly require VLMs to integrate information across multiple visual perspectives to build accurate world-centric 3D understanding. MultiView-Bench is designed to stress-test precisely these multi-view spatial reasoning capabilities, serving as a diagnostic benchmark before deploying VLMs in real-world 3D applications.
\end{tcolorbox}

\section{MultiView-Bench}

\subsection{Overview}
We introduce \textbf{MultiView-Bench}, a diagnostic benchmark designed to evaluate VLMs' ability to integrate multi-view observations into coherent, world-centric 3D spatial reasoning. As established in the motivation, real-world tasks like furniture assembly demand that models build accurate 3D mental representations from visual observations. Our benchmark is specifically constructed to stress-test these capabilities through three key design principles:

\begin{enumerate}
\item \textbf{World-centric coordinate system:} Real-world 3D applications (CAD software, robotics, manufacturing) operate in fixed global coordinate frames. We embed a visible coordinate system (X, Y, Z axes) that remains invariant across viewpoints, requiring models to decouple object positions from camera perspectives---the same allocentric reasoning needed to place furniture parts at precise global coordinates.

\item \textbf{Multi-view integration:} As illustrated in Figure~\ref{fig:furniture_example}, single-view observations often lead to visual misconceptions. Our benchmark requires models to synthesize information from multiple viewpoints to infer 3D spatial relationships that cannot be determined from any single view alone, directly testing the multi-view perception capability essential for assembly tasks.

\item \textbf{Controlled complexity:} To diagnose \emph{where} models fail, we design task variants with different degrees of freedom (DoF=1, 2, 3). This controlled progression from 1D to full 3D reasoning isolates specific failure modes, revealing whether models struggle with multi-view integration, axis identification, or full 3D spatial translation.
\end{enumerate}

Figure~\ref{fig:data_generation_pipeline} illustrates our modular data generation pipeline, which procedurally creates diverse 3D scenes with controlled variations. We describe each component below.


\subsection{Task Formulation}
The core task assesses a VLM's ability to reason about the relative positions of objects within a 3D scene. Given multiple viewpoint images, VLMs must describe a target object's position relative to a central object along each axis using the format \texttt{(±X/0, ±Y/0, ±Z/0)}, ensuring precise and parsable responses for large-scale evaluation.

To establish a viewpoint-independent frame of reference, we introduce a fixed global coordinate system with clearly marked axes---X (red), Y (green), and Z (blue). This coordinate system is implemented as fixed 3D meshes in Blender, remaining invariant across camera changes. We adjust object material opacity to ensure axes remain clearly visible. This setup mimics the Blender GUI, making MultiView-Bench a fundamental test for VLM-based agents intended for 3D modeling applications.

\subsection{Data Generation Pipeline}

We design a modular pipeline that procedurally generates diverse 3D scenes with controlled variations (Figure~\ref{fig:data_generation_pipeline}), consisting of five sequential stages: \textbf{(i) 3D Assets.} For synthetic tasks, we sample fundamental geometric objects (cubes, spheres, cylinders, cones). For real-world objects (3D Real World), we use 3DCoMPaT++ \citep{slim20253dcompat++}, which spans thousands of objects across categories such as tables, chairs, and airplanes. All objects are rescaled to a common bounding box to minimize visual ambiguity in spatial comparisons, allowing relative positions to be inferred reliably from any edge or vertex. Distractor objects can optionally be added to increase scene complexity. The pipeline is modular and easily extended to alternative 3D assets for domain-specific datasets. \textbf{(ii) Object Placement.} A central object is fixed at the origin, and other objects are randomly placed within a set minimum/maximum separation distance to avoid overlap while keeping them proximate. Deviations below a threshold margin along an axis are snapped to 0 to define exact relations. We also construct controlled layout variants to probe model limitations: DoF=1 (objects share the central object's axis), DoF=2 (objects share its plane), and DoF=3 (unconstrained placement). \textbf{(iii) Camera Viewpoints.} The main benchmark renders \textbf{six viewpoints with uniformly distributed azimuth angles and slight elevation}, ensuring all three axes remain visible. The pipeline supports arbitrary viewpoint configurations, enabling studies of VLM inductive biases and specialized tasks (e.g., egocentric-to-allocentric transformations, spatial navigation), as illustrated by extensions of \cite{jia2025omnispatial} and \cite{yin2025spatial} in Appendix \ref{more_spatial_tasks}. \textbf{(iv) Rendering.} Images are rendered per viewpoint; since this is parallelizable across CPU cores, large-scale dataset generation remains efficient and scalable despite the cost of complex, textured scenes. \textbf{(v) Q\&A Generation.} Question--answer pairs are generated automatically, with relative spatial relations computed directly from Blender's intrinsic coordinate system to ensure reliable ground-truth supervision.

\paragraph{Features of MultiView-Bench.}
Our pipeline offers several key advantages over manually curated benchmarks:

\begin{itemize}
\item \textbf{Principled construction:} Each design choice---world-centric coordinates, controlled DoF variants, multi-view requirements---directly targets a specific capability needed for real-world 3D reasoning. This principled approach ensures the benchmark measures what it claims to measure.

\item \textbf{Generalizability to arbitrary shapes:} The pipeline is agnostic to the 3D assets used. We demonstrate this by instantiating tasks with both geometric primitives and complex real-world objects from 3DCoMPaT++. Researchers can substitute domain-specific assets (e.g., mechanical parts, medical scans, architectural models) without modifying the core pipeline.

\item \textbf{Scalable task generation:} Unlike manually annotated datasets, our procedural approach can generate essentially unlimited task instances with guaranteed ground-truth labels. The main benchmark comprises 5 task variants with 100 instances each, plus 20 additional controlled variants for bias analysis---all generated automatically.

\item \textbf{Diagnostic granularity:} The DoF-controlled variants enable systematic ablation: by comparing performance across DoF=1, 2, and 3, we can isolate whether failures stem from 1D reasoning, 2D-to-3D lifting, or multi-view integration. This diagnostic power is essential for guiding future model improvements.
\end{itemize}

\section{Frontier Models Cannot Solve It Yet}

We evaluate 15 VLMs from the Claude, Gemini, and GPT families on MultiView-Bench. The run settings are reported in Appendix \ref{appendix:reproducibility}. Complete 15-model comparisons are provided Appendix \ref{full_table}.

\begin{figure*}[!t]
    \centering
    \includegraphics[width=\textwidth]{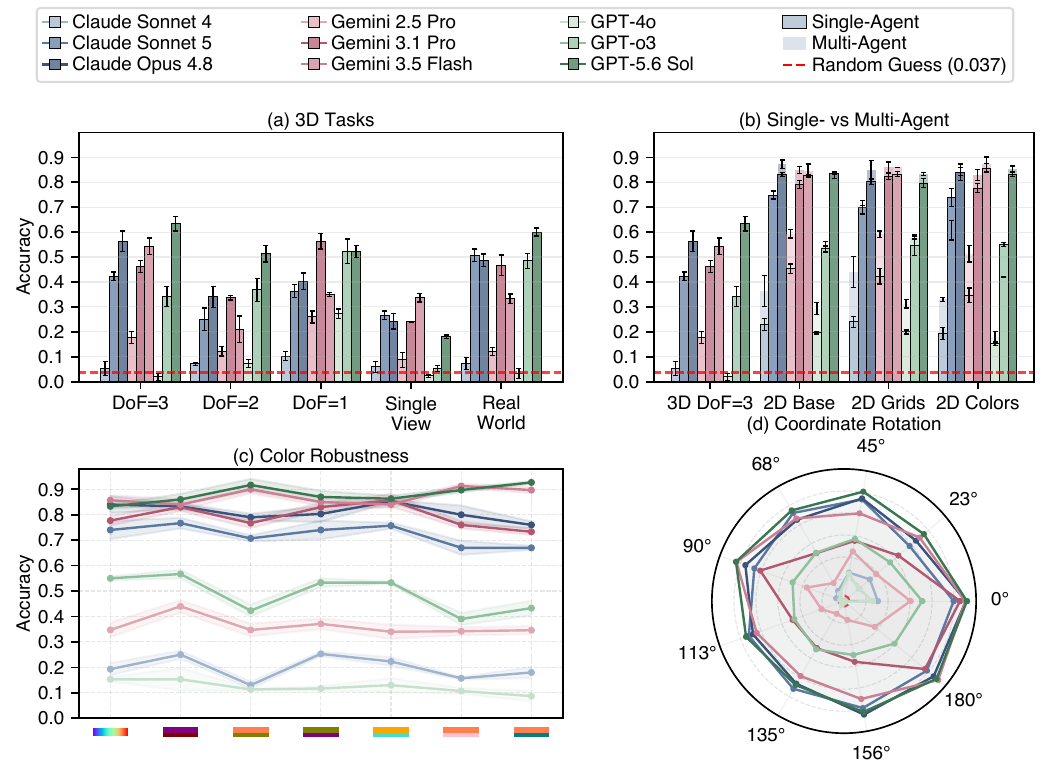}
    \caption{Representative results for three models from each provider family. \textbf{(a)} Accuracy across five 3D diagnostic settings. \textbf{(b)} A single-agent 3D DoF$=3$ reference and single-agent (opaque, bordered) versus view-wise multi-agent (lighter) accuracy on three 2D variants. \textbf{(c)} Accuracy across seven color-scheme conditions. \textbf{(d)} Accuracy under coordinate-axis rotations. Error bars and shaded bands show $\pm 1$ standard deviation; red dashed guides mark random chance ($1/27 \approx 3.7\%$). Complete 15-model comparisons are in Appendix \ref{full_table}. Hyperparameters and prompts used in experiments are in Appendix \ref{appendix:reproducibility}}
    \label{fig:MultiView-Bench_performance}
\end{figure*}

\begin{figure}
    \centering
\includegraphics[width=\textwidth]{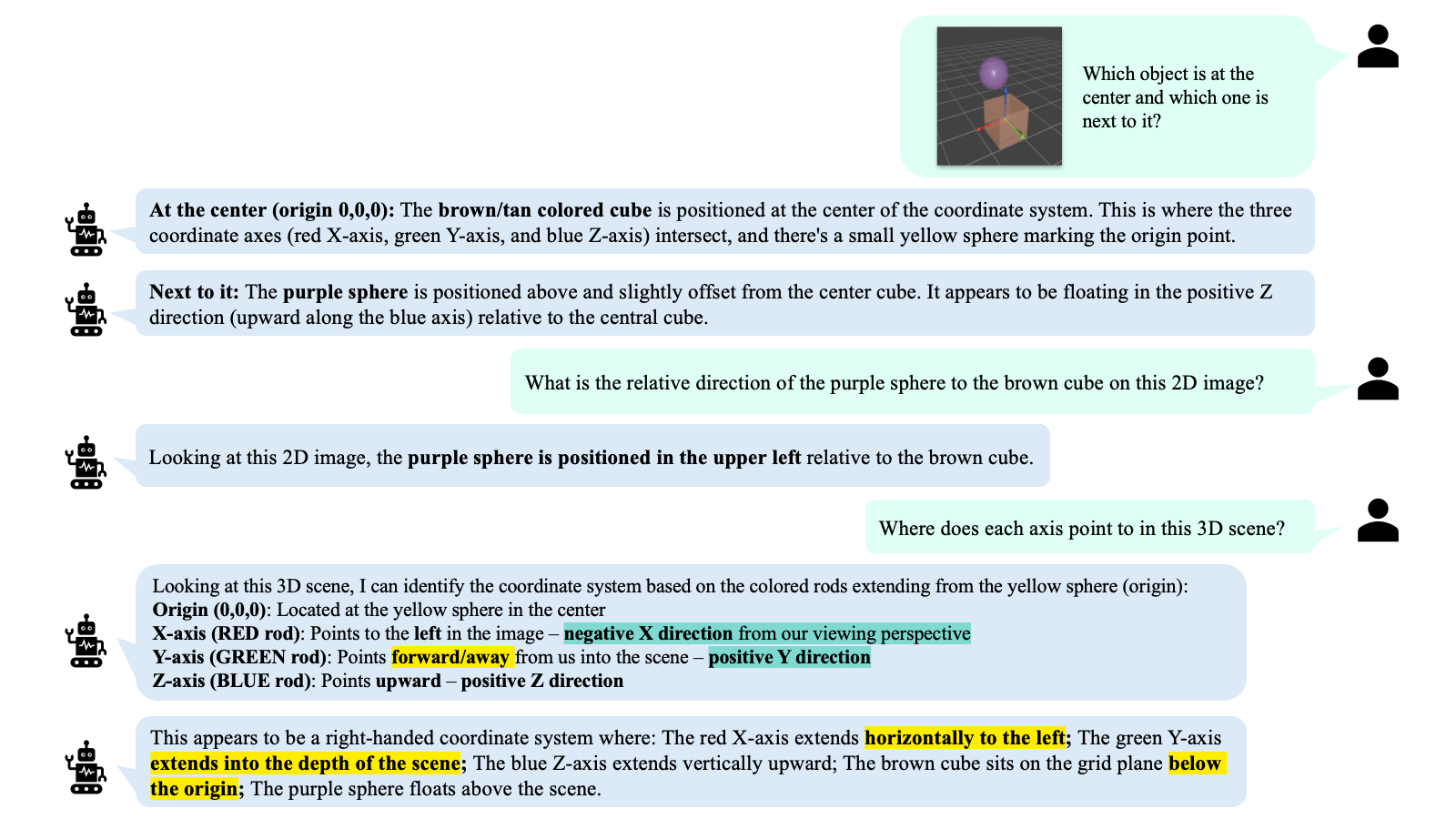}
\caption{Example conversation with VLM: VLM does well in 2D reasoning in the first 2 steps but fails to reason about the 3D scene in Step 3. Wrong reasoning traces are highlighted in yellow. Signs of inductive bias on the 3D coordinate system are highlighted in cyan.}
\label{fig:chat_example}
\end{figure}

\subsection{Performance Overview}

Figure~\ref{fig:MultiView-Bench_performance} reveals a striking finding: \highlight{most frontier models perform near random chance on the hardest tasks}. On 3D DoF=3 and 3D Real World, models like Claude series and GPT-4o barely exceed the \textbf{3.7\%} random baseline. Even GPT-5, the best-performing model, achieves only \textbf{50\%} accuracy on 3D DoF=3---far from reliable for real-world deployment. We have the following key observations:
\begin{itemize}
\item \textbf{Task difficulty scales with DoF:} As we decrease DoF from 3 to 1, accuracy improves dramatically. Claude 3.7 Sonnet reaches near GPT-5 level when DoF=1, suggesting models handle simpler spatial configurations but fail at full 3D reasoning.
\item \textbf{Single view is insufficient:} Providing only a single isometric view (clearly displaying all three axes) drops all models to random chance, confirming that \emph{multi-view integration is essential}---not merely helpful.
\item \textbf{Real-world objects are harder:} Performance on 3D Real World (using 3DCoMPaT++ assets) is consistently lower than synthetic primitives, indicating additional challenges from complex geometry and textures.
\end{itemize}

\subsection{Failure Pattern Analysis}

To understand \emph{why} models fail, we decompose the spatial reasoning task into intermediate 4 sub-steps analogous to human reasoning: \textbf{(1) Object Identification} -- distinguish individual objects; \textbf{(2) 2D Spatial Relation} -- articulate relative positions within a 2D viewpoint; \textbf{(3) Axis Direction Identification} -- interpret axis directionality in the 3D scene; and \textbf{(4) 3D Position Translation} -- translate 2D relations into 3D positional descriptions.







Our analysis reveals that VLMs succeed at Steps 1 and 2 but \highlight{fail critically at Step 3}---identifying axis directions in 3D context (Figure~\ref{fig:chat_example}). Models consistently struggle to express axis directions using unambiguous 3D spatial language, even when the coordinate system is clearly visible.

Motivated by this observation, we explore whether decomposing the task into simpler 2D views can improve performance. We design a 2D variant using three canonical views (front, side, top), each showing one 2D plane (XZ, YZ, XY) with only two axes per view (Figure \ref{fig:more_task_example}a). We test two configurations: (1) \textbf{Single-agent}: all three views provided simultaneously; (2) \textbf{Multi-agent}: each agent independently assesses one view, with final answers integrated.

Results (Figure~\ref{fig:MultiView-Bench_performance}b) reveal two key findings: \emph{(i) simplifying 3D to 2D dramatically improves performance}, and \emph{(ii) decomposing multi-view into single-view sub-tasks yields further gains depending on the models}. The benefit of view-wise multi-agent decomposition depends strongly on the capability of the base model. Across the seven older-generation models, decomposition increases mean 2D accuracy from 37.0\% to 46.3\%. In contrast, the eight newer-generation models already achieve 77.5\% in the single-agent setting, and decomposition barely changes their mean accuracy. This confirms that \highlight{older-generation VLMs struggle with both 3D perception and multi-view integration} while more recent models can already combine the three planar views effectively within a single context but still struggle with 3D perception.

\subsection{Bias Analysis}
\textbf{Visual Enhancements.} We explore how visual aids (color schemes, grids) influence performance. Prior work suggests structured visual aids improve VLM reasoning \citep{izadi2025visual}. Our experiments reveal model-specific effects: \highlight{grids and distinct colors help Claude series but \emph{surprisingly hurt Gemini 2.5 Pro}} (Figure \ref{fig:MultiView-Bench_performance}b). We also observe color biases---models show preferential responses to particular color combinations that diverge from human perception (Figure \ref{fig:MultiView-Bench_performance}c).

\textbf{Coordinate Rotation.} We identify a striking pattern: models frequently \emph{ignore explicitly depicted coordinate directions}, defaulting to the conventional right-handed coordinate system (Figure \ref{fig:chat_example}). To test this, we rotate axes to non-standard angles (23\textdegree{}, 45\textdegree{}, 68\textdegree{}, etc.). Results (Figure \ref{fig:MultiView-Bench_performance}d) confirm \emph{severe performance degradation under unconventional orientations}, revealing that VLMs heavily rely on learned coordinate-system priors rather than reasoning from visual evidence. This poses a fundamental challenge for deploying VLMs in real-world 3D applications where coordinate systems may not follow textbook conventions.

\begin{tcolorbox}[colback=MorandiBlue!5!white, colframe=MorandiBlue!80!black, title=\textbf{Key Takeaways}, boxsep=1mm, top=1mm, bottom=1mm]
\begin{itemize}
    \item Frontier VLMs can't solve 3D tasks (DoF=3), with many models under 50\% accuracy.
    \item Primary failure: \textbf{axis direction identification}---models succeed at 2D but fail at 3D.
    \item VLMs struggle with \textbf{both} 3D perception \textbf{and} multi-view integration.
    \item Strong \textbf{inductive biases} toward conventional coordinate systems.
\end{itemize}
\end{tcolorbox}

\section{ViewNavigator: Can We Augment VLMs with Agentic Reasoning?}

Given that frontier VLMs struggle with multi-view 3D spatial reasoning, a natural question arises: \highlight{can we augment their capabilities by embedding them in a multi-agent system?} We introduce \textbf{ViewNavigator}, a framework that combines VLMs and LLMs in a closed-loop architecture to enhance 3D spatial understanding---without requiring post-training or external geometry-based analysis.

\begin{figure*}[!t]
\begin{subfigure}[b]{0.48\textwidth}
        \centering
        \includegraphics[width=\textwidth]{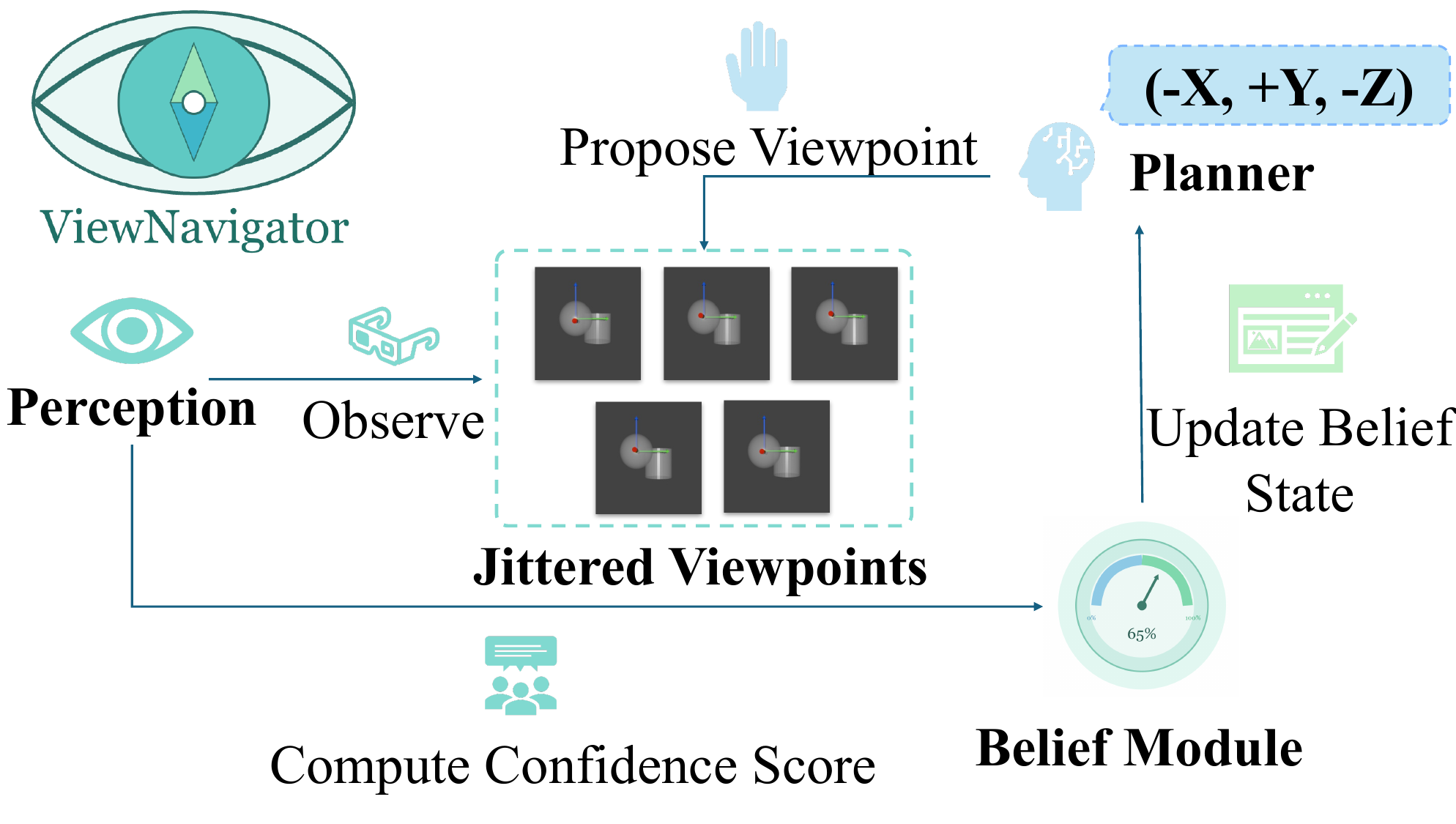}
        \caption{}
    \end{subfigure}
    \hfill
    \begin{subfigure}[b]{0.4\textwidth}
        \centering
        \includegraphics[width=\textwidth]{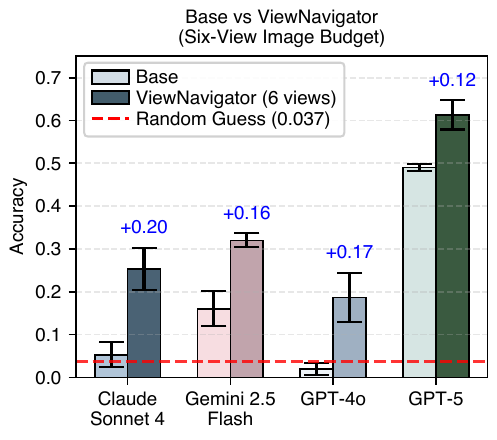}
        \caption{}
    \end{subfigure}
    \vspace{-2mm}
    \caption{(a) ViewNavigator workflow: LLM planner selects viewpoints, VLM provides judgments from micro-jittered views, belief module aggregates evidence via Dirichlet distribution. (b) \textbf{Budget-matched} ViewNavigator, constrained to the same budget as the base models ($\le 6$ viewpoints, no micro-jitter, uniform belief-updating), still improves all base models on 3D DoF$=$3; bars are means over three runs and error bars denote $\pm 1$ standard deviation. The extended-budget agent is reported in Appendix~\ref{sec:full_agent}.}
    \label{fig:ViewNavigator}
\end{figure*}

\subsection{Architecture}

ViewNavigator addresses VLM failures through three complementary components (Figure~\ref{fig:ViewNavigator}a):

\begin{itemize}
\item \textbf{Belief Module:} Since individual VLM predictions are noisy, we maintain a probabilistic belief over spatial directions using a Dirichlet distribution for each axis. By querying the VLM from multiple micro-jittered viewpoints (small perturbations around a base position), we aggregate votes and filter out unstable predictions through confidence-weighted updates.

\item \textbf{Active View Selection:} Rather than using fixed viewpoints, an LLM planner strategically selects the next viewpoint based on current uncertainty, using the current belief state and view history to gather maximally informative observations.

\item \textbf{Confidence-Gated Output:} The agent continues exploring viewpoints until the posterior probability on each axis exceeds a confidence threshold, preventing premature answers based on insufficient evidence while terminating quickly on easier cases.
\end{itemize}

\subsection{Workflow}

The closed-loop workflow operates iteratively (Figure~\ref{fig:ViewNavigator}a): (1) the LLM examines the current belief state and selects a viewpoint likely to reduce uncertainty; (2) the VLM observes the scene from multiple micro-jittered positions and provides categorical judgments for each axis; (3) the belief module aggregates these votes, discounting inconsistent predictions via confidence scoring; (4) the updated belief is checked against stopping criteria. This loop continues until the agent is sufficiently confident or reaches a maximum number of steps.

Individual VLM predictions are noisy, but aggregating them across viewpoints yields robust spatial judgments; full mathematical details are in Appendix~\ref{appendix:viewnavigator_details}.

\subsection{Results and Analysis}

To ensure that any gains reflect better reasoning rather than a larger test-time budget, we deploy ViewNavigator in a \textbf{budget-matched} configuration: the agent may propose at most six viewpoints in total---the same number of views provided to the base models---with micro-jitter disabled and uniform belief-updating, so that the number of images and the test-time compute exactly match the fixed-six-view baseline. Even under this strict budget, ViewNavigator \highlight{substantially improves all base models} on 3D DoF$=$3 (Figure~\ref{fig:ViewNavigator}b).

\begin{itemize}
\item \textbf{Weaker models benefit most:} GPT-4o improves from 2\% to 19\%, Claude 4 Sonnet from 5\% to 25\%. The agentic framework effectively compensates for base model limitations.
\item \textbf{Stronger models also improve:} GPT-5 improves from 49\% to 61\% , showing that even the best models benefit from structured multi-view reasoning.
\item \textbf{Diminishing returns:} Relative improvement decreases as base performance increases, suggesting a ceiling effect where agentic scaffolding cannot fully overcome architectural limitations.
\end{itemize}

Relaxing the budget constraint further amplifies these gains for GPT-5: we let ViewNavigator actively selecting up to 10 viewpoints, each with micro-jittered observations. It improves GPT-5's 3D DoF$=3$ accuracy by 27 percentage points, higher than budget-constrained case. However, we observe that letting weaker models select more viewpoints actually confuses them, resulting to worse performances than the 6-view case. We report the result of this full configuration in Appendix~\ref{sec:full_agent}.

\section{Conclusion}
\textbf{MultiView-Bench} diagnoses VLMs' ability to integrate multi-view observations into world-centric 3D understanding. All 15 frontier models remain unreliable on challenging 3D tasks, with failures concentrated in axis-direction identification and multi-view integration. \textbf{ViewNavigator}'s belief aggregation and active view selection improve all four evaluated base models even under a budget-matched setting. Together, they expose deployment-relevant limitations for spatial agents.

     \bibliography{references}

}{}

\ifthenelse{\equal{\compileversion}{bodyappendix}}{%
    
    \begin{tcolorbox}[
        colback=YaleBlue!4,
        colframe=white,
        arc=3mm,
        boxrule=1pt,
        top=4mm,
        bottom=4mm,
        left=4mm,
        right=4mm
    ]
        \begin{center}
            {\Large\bfseries MultiView-Bench: A Diagnostic Benchmark for World-Centric Multi-View Integration in VLMs\par}
            \vspace{0.5em}
            {\large Hantao Zhang$^{1,\dagger}$ \quad  Jinru Sui$^{2}$ \quad  Ed Li$^{1}$ \\ Dirk Bergemann$^{1}$ \quad  Zhuoran Yang$^{1}$\par}
            \vspace{0.5em}
            {\small
            $^1$Yale University \quad
            $^2$University of Edinburgh \par}
        \end{center}
        \vspace{0.5em}
        \noindent\textcolor{YaleBlue}{\textbf{Abstract.}}
        Recent benchmarks for VLMs largely assess single- or limited-view perception, leaving untested the core cognitive ability to \emph{integrate} observations across viewpoints into a coherent, world-centric (allocentric) 3D mental model. We introduce \textbf{MultiView-Bench}, a diagnostic benchmark expressly designed to evaluate multi-view integration for holistic 3D scene comprehension. Unlike existing datasets that focus on pixel-level mapping or camera-relative navigation, MultiView-Bench requires models to decouple object positioning from transient perspectives and ground them in a fixed global coordinate system. This capability serves as a \emph{prerequisite} for VLMs before being deployed for downstream tasks such as mechanical part assembly. Our systematic evaluation of frontier VLMs reveals consistent failure modes: strong performance on 2D planar relations from a single image, but marked difficulty with 3D spatial relations and with aggregating information across views. We further identify biases in VLMs, such as struggles with unconventional axis directions and sensitivity to object colorways and texture variations. Acknowledging these limitations, we propose \textbf{ViewNavigator}, which uses active viewpoint selection and evidence fusion to improve four base models by 12.3--20.0 percentage points under a six-image cap matching the fixed-view baseline; budget-extended gains are model-dependent and reach 27 percentage points for GPT-5.

        \vspace{2em}
        
        \noindent\textcolor{YaleBlue}{\textbf{Project Webpage:}} \href{https://hantaozhangrichard.github.io/MultiView-Bench/}{hantaozhangrichard.github.io/MultiView-Bench}

    \end{tcolorbox}

\begin{figure}[h]
    \centering
    \includegraphics[width=\linewidth]{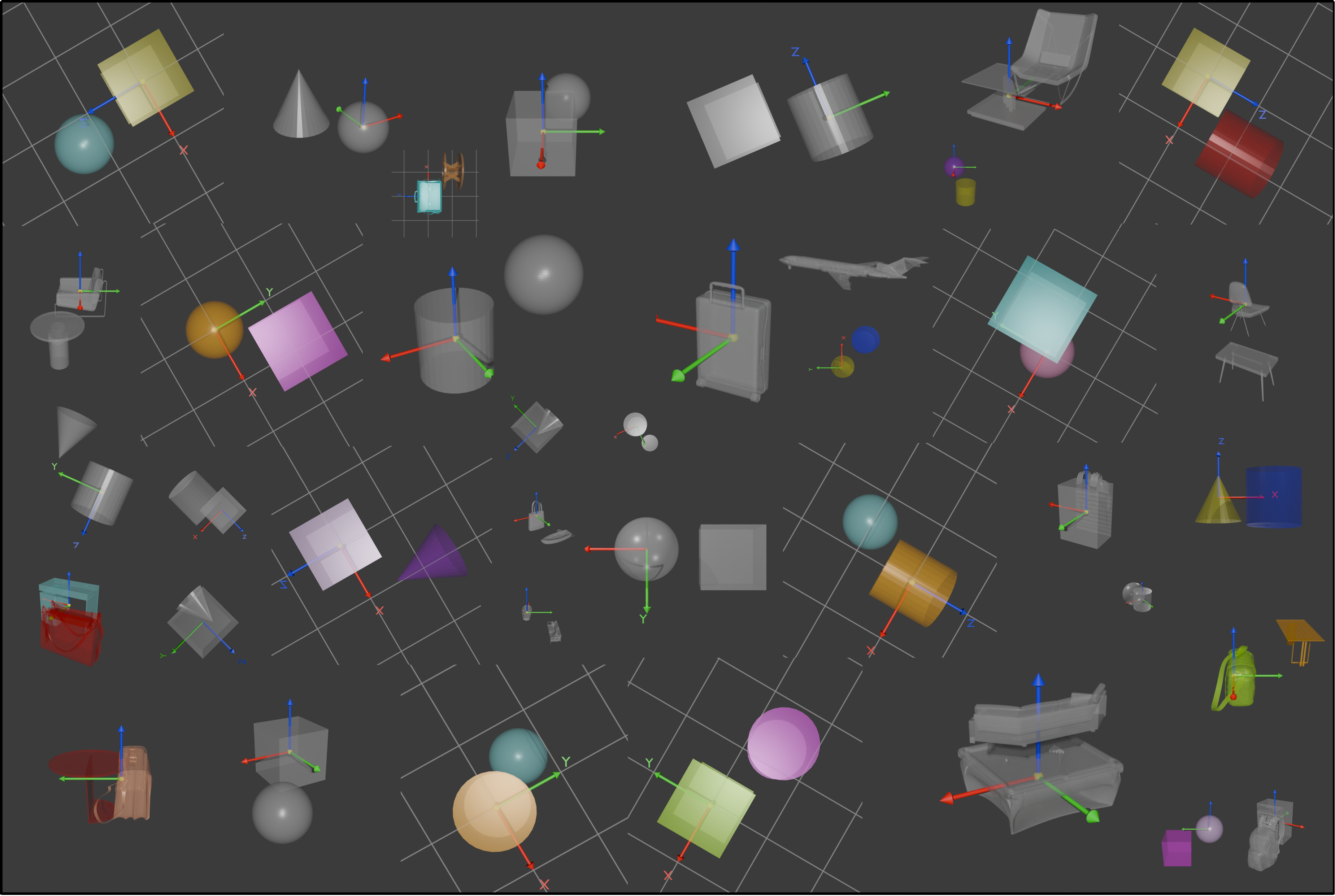}
    
    \label{fig:front_page}
\end{figure}
    
    
    \newpage
    \bibliography{references}
    
    \clearpage
    \appendix
\onecolumn

\setcounter{secnumdepth}{3}

\section*{\LARGE Appendices}
\label{sec:appendix}


\section{Evaluation Results}
\label{full_table}

\begingroup
\small
\begin{table*}[!ht]
  \centering
  \setlength{\tabcolsep}{3pt}
  \begin{tabular}{p{0.3\textwidth}rrrrr}
  \toprule
  Task & \shortstack{Claude\\Sonnet 3.7} & \shortstack{Claude\\Sonnet 4} & \shortstack{Claude\\Haiku 4.5} & \shortstack{Claude\\Sonnet 5} & \shortstack{Claude\\Opus 4.8} \\
  \midrule
  3D DoF=3 & $0.060$ & $0.053 \pm 0.029$ & $0.190 \pm 0.014$ & $0.423 \pm 0.017$ & $0.563 \pm 0.042$ \\
  3D DoF=2 & $0.150$ & $0.072 \pm 0.006$ & $0.110 \pm 0.022$ & $0.250 \pm 0.045$ & $0.340 \pm 0.043$ \\
  3D DoF=1 & $0.420$ & $0.103 \pm 0.017$ & $0.170 \pm 0.022$ & $0.363 \pm 0.026$ & $0.403 \pm 0.033$ \\
  3D Single View & $0.010$ & $0.060 \pm 0.022$ & $0.087 \pm 0.017$ & $0.267 \pm 0.017$ & $0.243 \pm 0.031$ \\
  3D Real World & $0.060$ & $0.073 \pm 0.025$ & $0.147 \pm 0.038$ & $0.507 \pm 0.025$ & $0.487 \pm 0.025$ \\
  2D 3 Views (Rotation 0) & $0.250$ & $0.230 \pm 0.024$ & $0.557 \pm 0.021$ & $0.747 \pm 0.017$ & $0.830 \pm 0.008$ \\
  2D 3 Views MA (Rot. 0) & $0.390$ & $0.363 \pm 0.062$ & $0.390 \pm 0.049$ & $0.750 \pm 0.016$ & $\boldsymbol{0.873 \pm 0.017}$ \\
  2D 3 Views Grids & $0.300$ & $0.240 \pm 0.022$ & $0.510 \pm 0.022$ & $0.700 \pm 0.028$ & $0.803 \pm 0.012$ \\
  2D 3 Views Grids multiagent & $0.400$ & $0.440 \pm 0.062$ & $0.393 \pm 0.025$ & $0.690 \pm 0.016$ & $0.850 \pm 0.037$ \\
  Colors: Random & $0.150$ & $0.193 \pm 0.024$ & $0.513 \pm 0.033$ & $0.740 \pm 0.036$ & $0.840 \pm 0.033$ \\
  Colors-MA: Random & $0.370$ & $0.330 \pm 0.008$ & $0.367 \pm 0.021$ & $0.607 \pm 0.039$ & $0.843 \pm 0.017$ \\
  Colors: Maroon+Purple & $0.190$ & $0.250 \pm 0.016$ & $0.583 \pm 0.026$ & $0.767 \pm 0.021$ & $0.833 \pm 0.012$ \\
  Colors: Turq.+Orange & $0.210$ & $0.223 \pm 0.017$ & $0.620 \pm 0.024$ & $0.757 \pm 0.012$ & $0.853 \pm 0.005$ \\
  Colors: Purple+Olive & $0.230$ & $0.253 \pm 0.009$ & $0.513 \pm 0.038$ & $0.740 \pm 0.037$ & $0.803 \pm 0.031$ \\
  Colors: Teal+Coral & $0.150$ & $0.180 \pm 0.022$ & $0.467 \pm 0.012$ & $0.670 \pm 0.008$ & $0.760 \pm 0.014$ \\
  Colors: Olive+Coral & $0.200$ & $0.130 \pm 0.014$ & $0.477 \pm 0.066$ & $0.707 \pm 0.012$ & $0.790 \pm 0.014$ \\
  Colors: Pink+Coral & $0.180$ & $0.157 \pm 0.005$ & $0.487 \pm 0.040$ & $0.670 \pm 0.028$ & $0.800 \pm 0.036$ \\
  2D 3 Views Rotation (23\textdegree{}) & $0.210$ & $0.230 \pm 0.022$ & $0.387 \pm 0.042$ & $0.583 \pm 0.029$ & $0.640 \pm 0.008$ \\
  2D 3 Views Rotation (45\textdegree{}) & $0.240$ & $0.197 \pm 0.012$ & $0.223 \pm 0.005$ & $0.703 \pm 0.025$ & $0.710 \pm 0.016$ \\
  2D 3 Views Rotation (68\textdegree{}) & $0.090$ & $0.080 \pm 0.033$ & $0.150 \pm 0.016$ & $0.693 \pm 0.052$ & $0.640 \pm 0.010$ \\
  2D 3 Views Rotation (90\textdegree{}) & $0.060$ & $0.057 \pm 0.005$ & $0.190 \pm 0.008$ & $0.650 \pm 0.014$ & $0.715 \pm 0.005$ \\
  2D 3 Views Rotation (113\textdegree{}) & $0.030$ & $0.017 \pm 0.005$ & $0.073 \pm 0.024$ & $0.690 \pm 0.024$ & $0.660 \pm 0.000$ \\
  2D 3 Views Rotation (135\textdegree{}) & $0.070$ & $0.000 \pm 0.000$ & $0.057 \pm 0.021$ & $\boldsymbol{0.690 \pm 0.036}$ & $0.657 \pm 0.025$ \\
  2D 3 Views Rotation (156\textdegree{}) & $0.030$ & $0.010 \pm 0.008$ & $0.113 \pm 0.005$ & $0.740 \pm 0.000$ & $\boldsymbol{0.785 \pm 0.035}$ \\
  2D 3 Views Rotation (180\textdegree{}) & $0.000$ & $0.000 \pm 0.000$ & $0.243 \pm0.031$ & $0.737 \pm 0.005$ & $0.797 \pm 0.009$ \\
  \bottomrule
  \end{tabular}
  \caption{MultiView-Bench accuracy results for the Claude model family. Cells shown as $x \pm s$ report the mean and standard deviation over 3 runs; a single value denotes one run. Bold indicates the highest mean in each task across all 15 evaluated models.}
  \label{tab:full_results_claude}
\end{table*}

\begin{table*}[!p]
  \centering
  \setlength{\tabcolsep}{3pt}
  \begin{tabular}{p{0.3\textwidth}rrrr}
  \toprule
  Task & 2.5 Flash & 2.5 Pro & 3.5 Flash & 3.1 Pro \\
  \midrule
  3D DoF=3 & $0.160 \pm 0.041$ & $0.177 \pm 0.025$ & $0.543 \pm 0.034$ & $0.463 \pm 0.024$ \\
  3D DoF=2 & $0.120 \pm 0.022$ & $0.122 \pm 0.019$ & $0.210 \pm 0.054$ & $0.337 \pm 0.009$ \\
  3D DoF=1 & $0.217 \pm 0.005$ & $0.260 \pm 0.024$ & $0.350 \pm 0.008$ & $0.563 \pm 0.031$ \\
  3D Single View & $0.090 \pm 0.016$ & $0.088 \pm 0.030$ & $\boldsymbol{0.337 \pm 0.017}$ & $0.240 \pm 0.000$ \\
  3D Real World & $0.107 \pm 0.025$ & $0.120 \pm 0.016$ & $0.333 \pm 0.019$ & $0.467 \pm 0.041$ \\
  2D 3 Views (Rotation 0) & $0.340 \pm 0.014$ & $0.453 \pm 0.019$ & $0.827 \pm 0.005$ & $0.790 \pm 0.016$ \\
  2D 3 Views MA (Rot. 0) & $0.500 \pm 0.037$ & $0.593 \pm 0.017$ & $0.847 \pm 0.026$ & $0.850 \pm 0.014$ \\
  2D 3 Views Grids & $0.327 \pm 0.029$ & $0.423 \pm 0.031$ & $0.833 \pm 0.009$ & $0.823 \pm 0.012$ \\
  2D 3 Views Grids multiagent & $0.537 \pm 0.031$ & $0.592 \pm 0.013$ & $\boldsymbol{0.860 \pm 0.000}$ & $0.860 \pm 0.022$ \\
  Colors: Random & $0.343 \pm 0.019$ & $0.347 \pm 0.029$ & $\boldsymbol{0.857 \pm 0.019}$ & $0.777 \pm 0.019$ \\
  Colors-MA: Random & $0.440 \pm 0.014$ & $0.513 \pm 0.033$ & $\boldsymbol{0.873 \pm 0.029}$ & $0.830 \pm 0.022$ \\
  Colors: Maroon+Purple & $0.313 \pm 0.024$ & $0.440 \pm 0.024$ & $0.840 \pm 0.022$ & $0.830 \pm 0.016$ \\
  Colors: Turq.+Orange & $0.277 \pm 0.017$ & $0.340 \pm 0.028$ & $0.840 \pm 0.014$ & $0.857 \pm 0.025$ \\
  Colors: Purple+Olive & $0.317 \pm 0.021$ & $0.371 \pm 0.021$ & $0.850 \pm 0.016$ & $0.830 \pm 0.016$ \\
  Colors: Teal+Coral & $0.227 \pm 0.021$ & $0.346 \pm 0.016$ & $0.897 \pm 0.012$ & $0.733 \pm 0.012$ \\
  Colors: Olive+Coral & $0.210 \pm 0.014$ & $0.347 \pm 0.025$ & $0.900 \pm 0.016$ & $0.767 \pm 0.019$ \\
  Colors: Pink+Coral & $0.183 \pm 0.039$ & $0.342 \pm 0.016$ & $\boldsymbol{0.913 \pm 0.009}$ & $0.760 \pm 0.014$ \\
  2D 3 Views Rotation (23\textdegree{}) & $0.310 \pm 0.016$ & $0.286 \pm 0.010$ & $0.673 \pm 0.037$ & $0.483 \pm 0.019$ \\
  2D 3 Views Rotation (45\textdegree{}) & $0.303 \pm 0.041$ & $0.344 \pm 0.005$ & $0.607 \pm 0.019$ & $0.417 \pm 0.009$ \\
  2D 3 Views Rotation (68\textdegree{}) & $0.260 \pm 0.042$ & $0.142 \pm 0.021$ & $0.650 \pm 0.036$ & $0.377 \pm 0.012$ \\
  2D 3 Views Rotation (90\textdegree{}) & $0.167 \pm 0.039$ & $0.270 \pm 0.016$ & $0.780 \pm 0.014$ & $0.607 \pm 0.026$ \\
  2D 3 Views Rotation (113\textdegree{}) & $0.223 \pm 0.040$ & $0.163 \pm 0.012$ & $0.633 \pm 0.012$ & $0.373 \pm 0.017$ \\
  2D 3 Views Rotation (135\textdegree{}) & $0.127 \pm 0.025$ & $0.100 \pm 0.024$ & $0.590 \pm 0.036$ & $0.370 \pm 0.008$ \\
  2D 3 Views Rotation (156\textdegree{}) & $0.130 \pm 0.043$ & $0.130 \pm 0.016$ & $0.677 \pm 0.019$ & $0.420 \pm 0.014$ \\
  2D 3 Views Rotation (180\textdegree{}) & $0.123 \pm 0.021$ & $0.273 \pm 0.021$ & $\boldsymbol{0.837 \pm 0.009}$ & $0.720 \pm 0.014$ \\
  \bottomrule
  \end{tabular}
  \caption{MultiView-Bench accuracy results for the Gemini model family. Cells shown as $x \pm s$ report the mean and standard deviation over 3 runs; a single value denotes one run. Bold indicates the highest mean in each task across all 15 evaluated models.}
  \label{tab:full_results_gemini}
\end{table*}

\begin{table*}[!p]
  \centering
  \setlength{\tabcolsep}{3pt}
  \resizebox{\textwidth}{!}{%
  \begin{tabular}{p{0.3\textwidth}rrrrrr}
  \toprule
  Task & 4o & o3 & 5 & 5.6 Sol & 5.6 Terra & 5.6 Luna \\
  \midrule
  3D DoF=3 & $0.020 \pm 0.014$ & $0.342 \pm 0.041$ & $0.490 \pm 0.008$ & $\boldsymbol{0.633 \pm 0.029}$ & $0.497 \pm 0.029$ & $0.457 \pm 0.021$ \\
  3D DoF=2 & $0.073 \pm 0.017$ & $0.368 \pm 0.045$ & $0.373 \pm 0.047$ & $\boldsymbol{0.513 \pm 0.033}$ & $0.370 \pm 0.008$ & $0.443 \pm 0.037$ \\
  3D DoF=1 & $0.273 \pm 0.019$ & $0.523 \pm 0.050$ & $0.397 \pm 0.025$ & $0.523 \pm 0.024$ & $0.427 \pm 0.012$ & $\boldsymbol{0.637 \pm 0.005}$ \\
  3D Single View & $0.023 \pm 0.005$ & $0.053 \pm 0.013$ & $0.127 \pm 0.017$ & $0.180 \pm 0.008$ & $0.227 \pm 0.009$ & $0.150 \pm 0.014$ \\
  3D Real World & $0.033 \pm 0.019$ & $0.485 \pm 0.030$ & $0.567 \pm 0.034$ & $\boldsymbol{0.600 \pm 0.016}$ & $0.460 \pm 0.033$ & $0.600 \pm 0.016$ \\
  2D 3 Views (Rotation 0) & $0.197 \pm 0.005$ & $0.533 \pm 0.012$ & $0.637 \pm 0.005$ & $\boldsymbol{0.837 \pm 0.005}$ & $0.813 \pm 0.029$ & $0.833 \pm 0.024$ \\
  2D 3 Views MA (Rot. 0) & $0.293 \pm 0.024$ & $0.540 \pm 0.022$ & $0.693 \pm 0.025$ & $0.827 \pm 0.012$ & $0.810 \pm 0.000$ & $0.773 \pm 0.025$ \\
  2D 3 Views Grids & $0.200 \pm 0.008$ & $0.547 \pm 0.041$ & $0.690 \pm 0.000$ & $0.797 \pm 0.017$ & $\boldsymbol{0.870 \pm 0.016}$ & $0.820 \pm 0.028$ \\
  2D 3 Views Grids multiagent & $0.313 \pm 0.017$ & $0.577 \pm 0.005$ & $0.660 \pm 0.059$ & $0.833 \pm 0.005$ & $0.827 \pm 0.005$ & $0.833 \pm 0.031$ \\
  Colors: Random & $0.153 \pm 0.009$ & $0.550 \pm 0.008$ & $0.663 \pm 0.017$ & $0.833 \pm 0.009$ & $0.827 \pm 0.026$ & $0.817 \pm 0.012$ \\
  Colors-MA: Random & $0.180 \pm 0.022$ & $0.420 \pm 0.000$ & $0.577 \pm 0.102$ & $0.853 \pm 0.012$ & $0.827 \pm 0.005$ & $0.830 \pm 0.014$ \\
  Colors: Maroon+Purple & $0.153 \pm 0.045$ & $0.567 \pm 0.017$ & $0.687 \pm 0.033$ & $\boldsymbol{0.860 \pm 0.022}$ & $0.843 \pm 0.033$ & $0.803 \pm 0.005$ \\
  Colors: Turq.+Orange & $0.130 \pm 0.028$ & $0.533 \pm 0.005$ & $0.723 \pm 0.012$ & $\boldsymbol{0.863 \pm 0.012}$ & $0.833 \pm 0.031$ & $0.800 \pm 0.037$ \\
  Colors: Purple+Olive & $0.117 \pm 0.009$ & $0.533 \pm 0.017$ & $0.660 \pm 0.000$ & $\boldsymbol{0.870 \pm 0.024}$ & $0.857 \pm 0.005$ & $0.790 \pm 0.022$ \\
  Colors: Teal+Coral & $0.087 \pm 0.021$ & $0.433 \pm 0.029$ & $0.580 \pm 0.037$ & $\boldsymbol{0.927 \pm 0.005}$ & $0.887 \pm 0.012$ & $0.867 \pm 0.019$ \\
  Colors: Olive+Coral & $0.113 \pm 0.005$ & $0.423 \pm 0.024$ & $0.610 \pm 0.022$ & $\boldsymbol{0.917 \pm 0.024}$ & $0.887 \pm 0.012$ & $0.873 \pm 0.012$ \\
  Colors: Pink+Coral & $0.107 \pm 0.012$ & $0.390 \pm 0.024$ & $0.587 \pm 0.025$ & $0.897 \pm 0.009$ & $0.887 \pm 0.033$ & $0.837 \pm 0.005$ \\
  2D 3 Views Rotation (23\textdegree{}) & $0.130 \pm 0.008$ & $0.410 \pm 0.029$ & $0.507 \pm 0.029$ & $\boldsymbol{0.710 \pm 0.008}$ & $0.670 \pm 0.037$ & $0.610 \pm 0.022$ \\
  2D 3 Views Rotation (45\textdegree{}) & $0.190 \pm 0.022$ & $0.430 \pm 0.029$ & $0.487 \pm 0.033$ & $\boldsymbol{0.757 \pm 0.021}$ & $0.677 \pm 0.005$ & $0.673 \pm 0.009$ \\
  2D 3 Views Rotation (68\textdegree{}) & $0.057 \pm 0.019$ & $0.380 \pm 0.008$ & $0.460 \pm 0.014$ & $\boldsymbol{0.713 \pm 0.039}$ & $0.603 \pm 0.026$ & $0.623 \pm 0.034$ \\
  2D 3 Views Rotation (90\textdegree{}) & $0.027 \pm 0.009$ & $0.373 \pm 0.029$ & $0.510 \pm 0.036$ & $0.783 \pm 0.012$ & $0.770 \pm 0.029$ & $\boldsymbol{0.787 \pm 0.019}$ \\
  2D 3 Views Rotation (113\textdegree{}) & $0.007 \pm 0.005$ & $0.360 \pm 0.029$ & $0.470 \pm 0.037$ & $\boldsymbol{0.710 \pm 0.014}$ & $0.570 \pm 0.033$ & $0.610 \pm 0.036$ \\
  2D 3 Views Rotation (135\textdegree{}) & $0.033 \pm 0.012$ & $0.380 \pm 0.024$ & $0.400 \pm 0.016$ & $0.660 \pm 0.008$ & $0.597 \pm 0.025$ & $0.543 \pm 0.038$ \\
  2D 3 Views Rotation (156\textdegree{}) & $0.003 \pm 0.005$ & $0.373 \pm 0.019$ & $0.473 \pm 0.012$ & $0.767 \pm 0.005$ & $0.657 \pm 0.021$ & $0.677 \pm 0.045$ \\
  2D 3 Views Rotation (180\textdegree{}) & $0.007 \pm 0.009$ & $0.450 \pm 0.008$ & $0.587 \pm 0.033$ & $0.823 \pm 0.017$ & $0.783 \pm 0.037$ & $0.803 \pm 0.025$ \\
  \bottomrule
  \end{tabular}
  }
  \caption{MultiView-Bench accuracy results for the GPT model family. Cells shown as $x \pm s$ report the mean and standard deviation over 3 runs; a single value denotes one run. Bold indicates the highest mean in each task across all 15 evaluated models.}
  \label{tab:full_results_gpt}
\end{table*}

\endgroup

\clearpage

\begin{figure*}[!t]
    \section{Complete Model Comparison Figures}
    This section provides the complete model-comparison plots corresponding to the main-paper analyses.
    \par\medskip
    \centering
    \includegraphics[width=\textwidth]{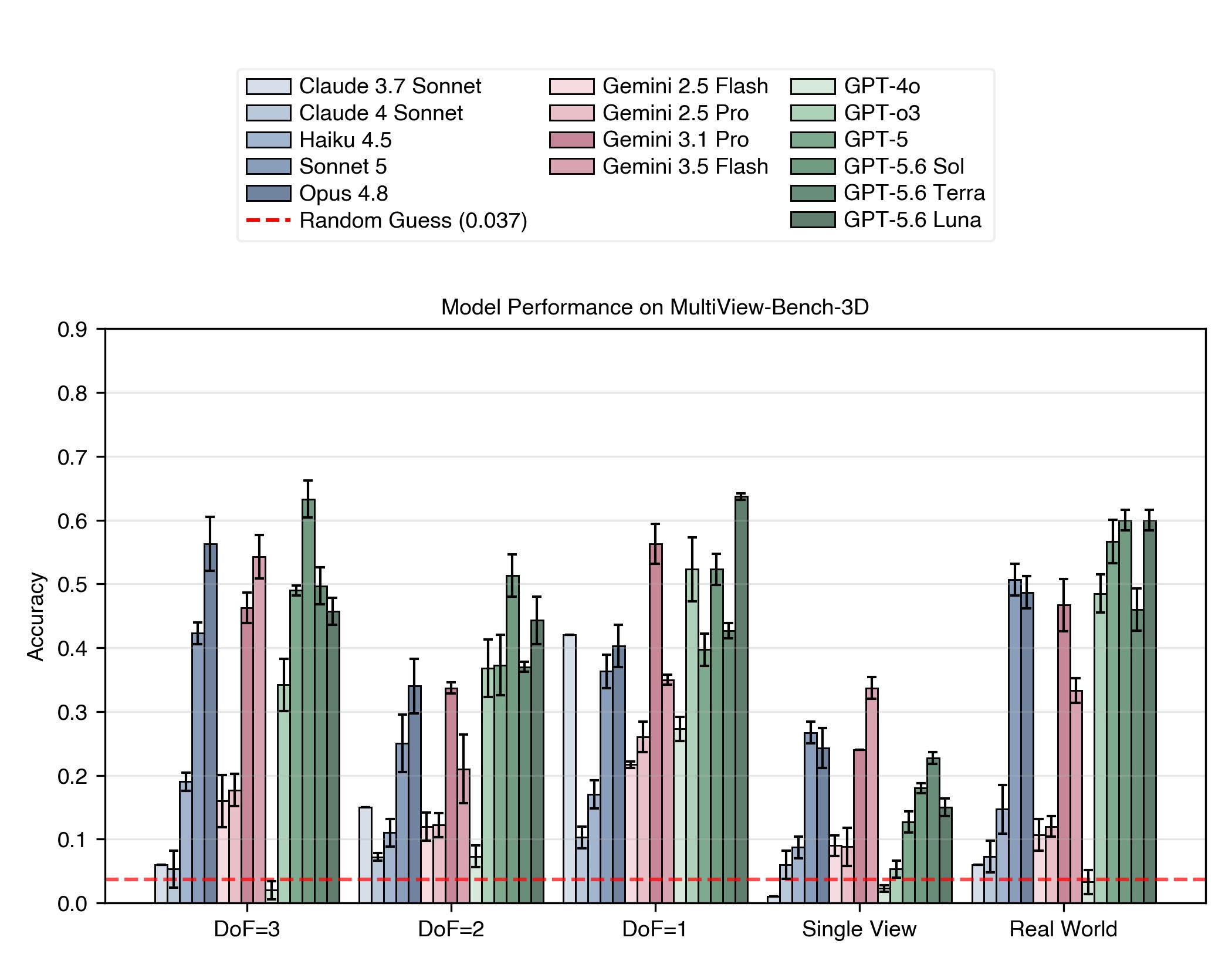}
    \caption{Complete 15-model accuracy comparison across the five MultiView-Bench 3D variants. Error bars show $\pm 1$ standard deviation where repeated runs are available, and the dashed line marks random chance.}
    \label{fig:full_3d_results}
\end{figure*}

\begin{figure*}[!p]
    \centering
    \includegraphics[width=\textwidth]{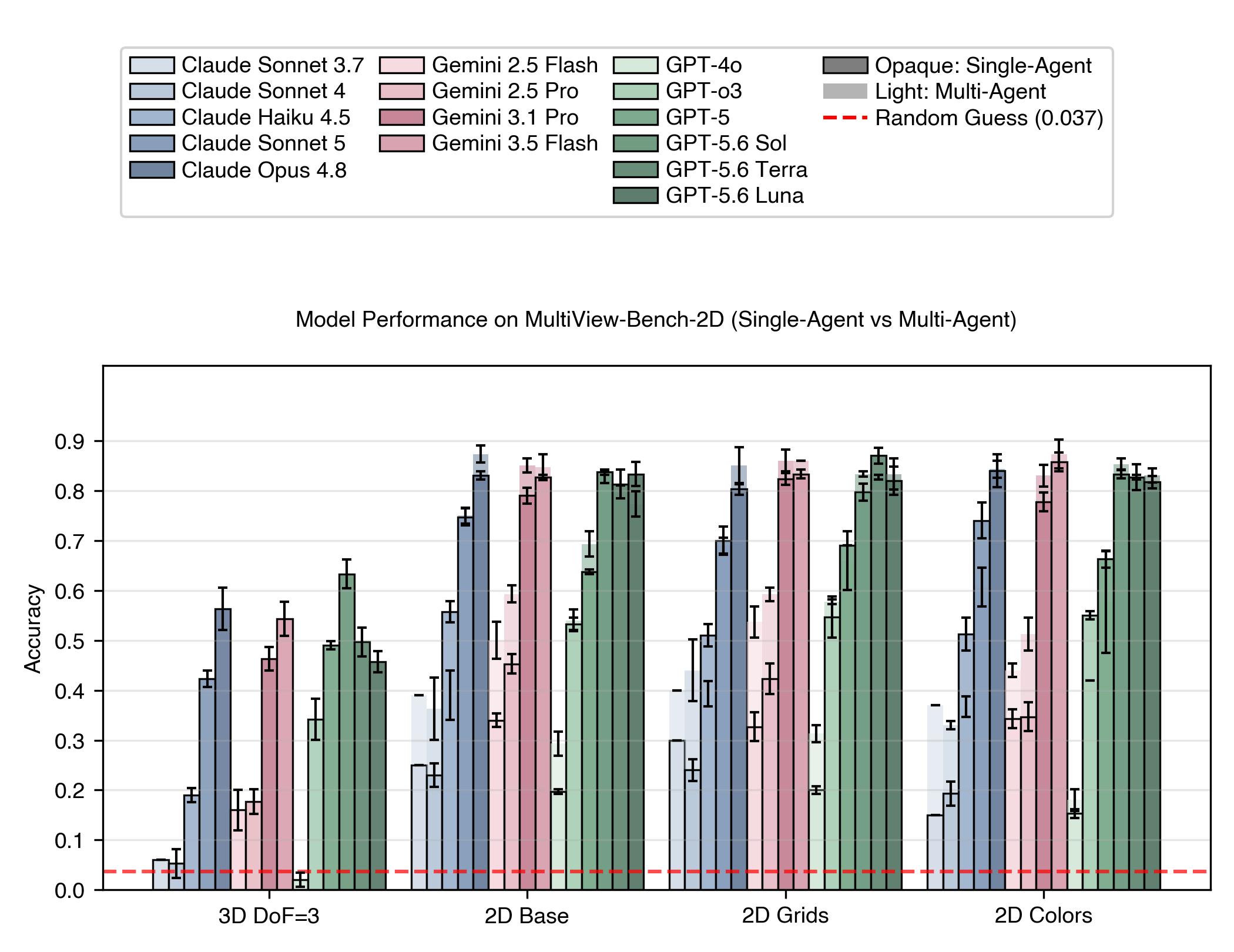}
    \caption{Complete 15-model single-agent versus multi-agent comparison. Opaque, bordered bars denote single-agent results, while lighter bars denote multi-agent results.}
    \label{fig:full_2d_results}
\end{figure*}

\begin{figure*}[!p]
    \centering
    \includegraphics[width=\textwidth]{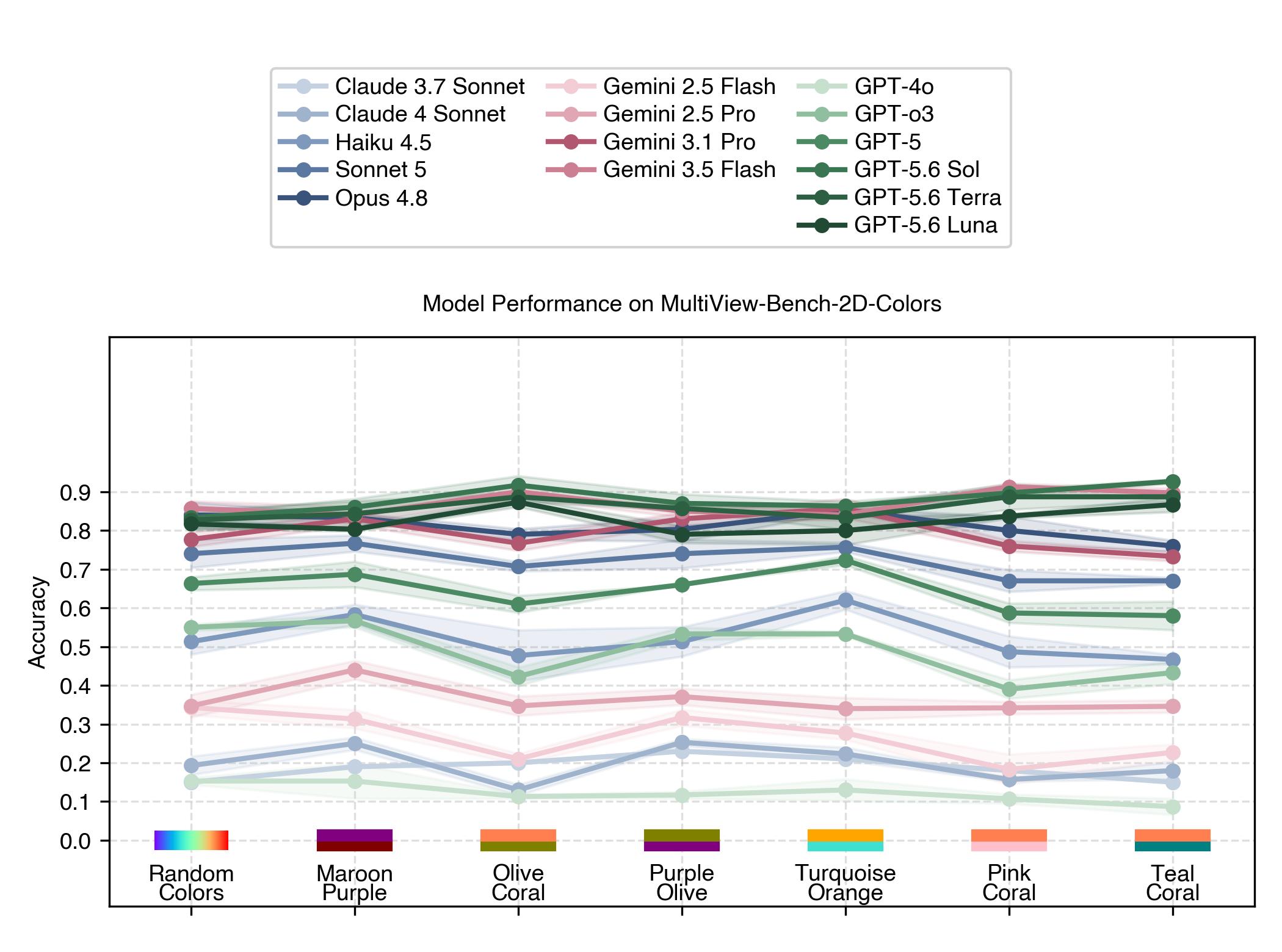}
    \caption{Complete 15-model color-scheme robustness comparison. Shaded bands show $\pm 1$ standard deviation.}
    \label{fig:full_color_results}
\end{figure*}

\begin{figure*}[!p]
    \centering
    \includegraphics[width=\textwidth]{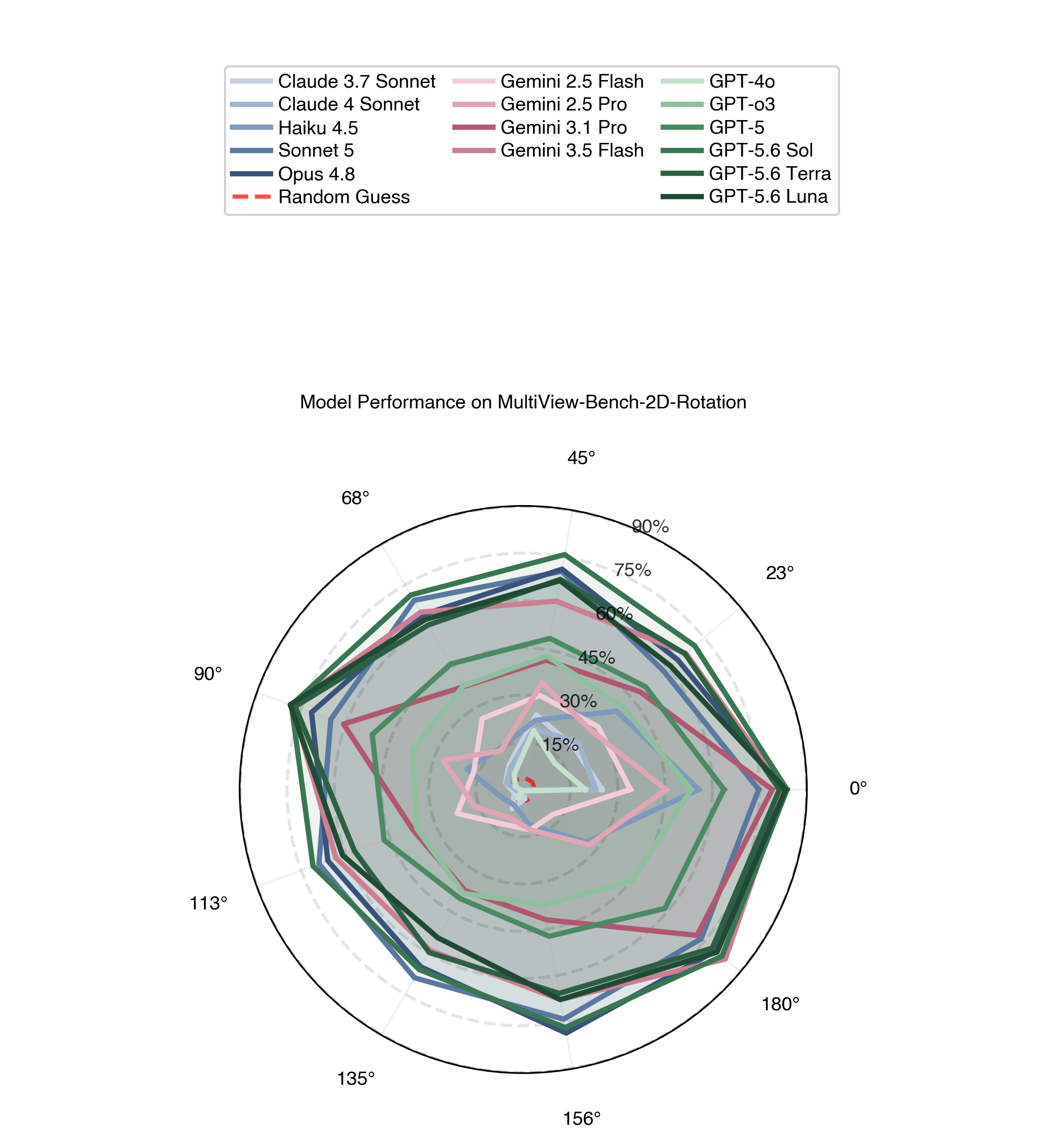}
    \caption{Complete 15-model coordinate-rotation robustness comparison. The radial scale is linear with a 90\% outer ring and a dashed random-chance ring.}
    \label{fig:full_rotation_results}
\end{figure*}
\clearpage

\section{More Example Tasks}

\begin{figure}[h]
\begin{center}
\includegraphics[width=\textwidth]{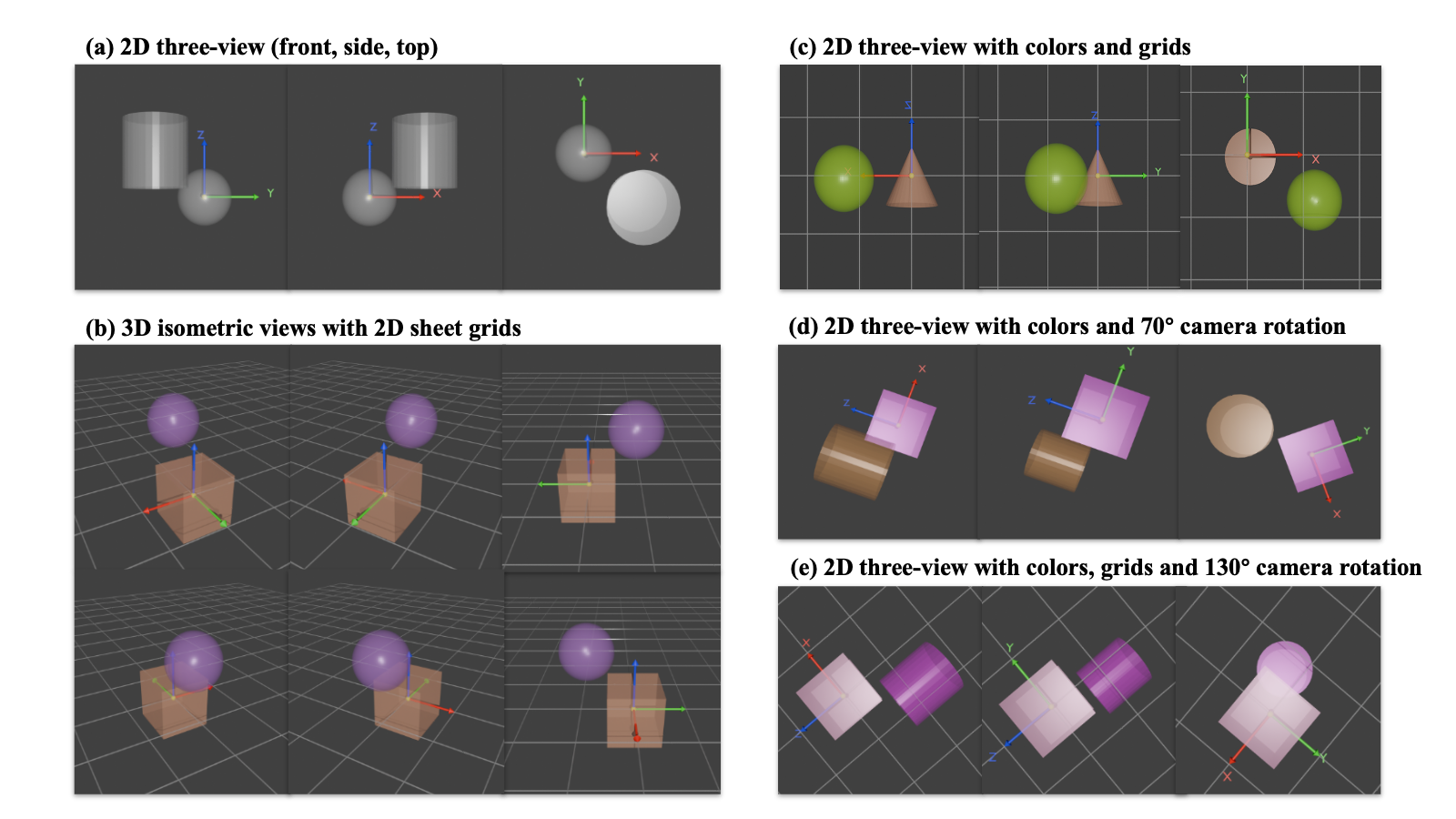}
\end{center}
\caption{More Example Tasks}
\label{fig:more_task_example}
\end{figure}

\newpage

\section{ViewNavigator: Technical Details}
\label{appendix:viewnavigator_details}

This section provides the full mathematical details of the ViewNavigator framework introduced in the main paper.

\subsection{Belief Module}

The agent maintains a probabilistic belief over spatial directions for each axis using a Dirichlet distribution parameterized by vector $\boldsymbol{\alpha}_A = [\alpha_{A,+}, \alpha_{A,0}, \alpha_{A,-}]$ for each axis $A \in \{X,Y,Z\}$. Initially, each axis is given a uniform prior: $\boldsymbol{\alpha}_A = [1,1,1]$.

\paragraph{Micro-Jittered Sampling.} Upon selecting a camera viewpoint, the agent captures multiple images using micro-jitters (small perturbations around a base viewpoint) to assess stability in the VLM's answers. For each jittered viewpoint, the VLM returns a categorical judgment $(+,0,-)$ independently for each axis, resulting in vote counts $k_{A,+}, k_{A,0}, k_{A,-}$ from a set of $n$ images.

\paragraph{Smoothed Proportions.} To update the belief, we first compute smoothed proportions:
\begin{equation}
\hat{p}_{A,s} = \frac{k_{A,s} + \lambda}{n + 3\lambda}, \quad s \in \{+,0,-\}
\end{equation}
where $\lambda$ is a smoothing constant (default $\lambda=1$) to avoid over-confidence for small $n$.

\subsection{Confidence Scoring}

To guard against unstable or noisy predictions, we discount the evidence size with a confidence score. We propose two methods:

\paragraph{Wilson Lower Bound Score.} Given that the majority label among the $n$ answers occurs $k_\text{max}$ times, the empirical majority proportion is $\hat{p} = k_\text{max}/n$. The Wilson score interval offers a conservative estimate of the true binomial proportion, particularly robust for small $n$ or when proportions are near 0 or 1. The 95\% Wilson lower bound is computed as:
\begin{equation}
\text{LB} = \frac{\hat{p} + z^2/(2n) - z\sqrt{\frac{\hat{p}(1 - \hat{p})}{n} + \frac{z^2}{4n^2}}}{1 + z^2/n}, \quad z=1.96.
\end{equation}
We map this to a conservative confidence score relative to a random baseline (uniform guess = $1/3$):
\begin{equation}
\omega_A = \left( \frac{\max(\text{LB},\, 1/3) - 1/3}{2/3} \right)^\gamma,
\end{equation}
where $\gamma \in [1, 2]$ controls sensitivity.

\paragraph{Relative Entropy Score.} Let $H(\hat{\mathbf{p}}_A) = -\sum_{s} \hat{p}_{A,s} \log \hat{p}_{A,s}$ be the entropy of the smoothed vote distribution and $H_{\max} = \log 3$ its maximum for three equally likely outcomes. The normalized entropy gap from uniform is:
\begin{equation}
\omega_A = \left( 1 - \frac{H(\hat{\mathbf{p}}_A)}{H_{\max}} \right)^\gamma,
\end{equation}
with $\gamma$ again controlling sensitivity. Both methods prevent overconfidence when a cluster's votes are unstable, rewarding highly peaked vote distributions and penalizing near-uniform ones.

\subsection{Belief Update}

The effective evidence size is:
\begin{equation}
n_{\text{eff},A} = n \cdot \omega_A.
\end{equation}

The smoothed proportions $\hat{p}_{A,s}$ are scaled by $n_{\text{eff},A}$ to yield soft counts:
\begin{equation}
\Delta\alpha_{A,s} = n_{\text{eff},A} \cdot \hat{p}_{A,s}, \quad \forall s \in \{+,0,-\}
\end{equation}
These are added to the Dirichlet parameters to yield the new belief:
\begin{equation}
\alpha_{A,s} \leftarrow \alpha_{A,s} + \Delta\alpha_{A,s}
\end{equation}
This belief is updated iteratively over successive jittered view clusters.

\subsection{Active View Selection and Stopping Criteria}

The LLM planner actively proposes the next best camera viewpoint based on the current belief state and previous view history, aiming to maximize information gain and reduce uncertainty. After each belief update, the agent checks if:
\begin{enumerate}
\item The posterior probability of the dominant class on each axis exceeds a confidence threshold $\tau$ (default $\tau=0.6$)
\item Sufficient evidence concentration is reached: total evidence $\sum_s \alpha_{A,s} \geq \kappa_{\text{min}}$
\end{enumerate}

If these criteria are met for all axes, the agent terminates the viewpoint exploration and outputs the final prediction:
\begin{equation}
\text{prediction} = \arg\max_{s \in \{+,0,-\}} \frac{\alpha_{A,s}}{\sum_t \alpha_{A,t}}, \quad \forall A \in \{X,Y,Z\}.
\end{equation}

\newpage

\begin{table*}[!t]
\section{Experiments Setup}
\label{appendix:reproducibility}

All experiments are conducted using our proposed \textbf{MultiView-Bench} benchmark. The 3D scenes and corresponding multi-view images are procedurally generated using Blender. Each scene is constructed with a fixed global coordinate system, represented by colored axes (X: red, Y: green, Z: blue), to provide a consistent frame of reference across all viewpoints. The generation pipeline allows for the use of various 3D assets, randomized object placements, and configurable camera positions, as detailed in the main paper.

\subsection{Model Reasoning Configuration}
Table~\ref{tab:model_reasoning_config} records the request-side reasoning control associated with every model column in the complete benchmark table. Seven request paths did not send a provider-specific reasoning or thinking field; these entries are marked \emph{default}. Thus, the table distinguishes an explicit request setting from the effective default mode. The other eight model columns use the explicit controls shown below.

\centering
\caption{Request-side reasoning configuration used for each reported model column.}
\label{tab:model_reasoning_config}
\small
\begin{tabular}{@{}p{0.18\textwidth}p{0.31\textwidth}p{0.45\textwidth}@{}}
\toprule
\textbf{Model} & \textbf{API model ID} & \textbf{Request field and effective reasoning mode} \\
\midrule
Claude 3.7 Sonnet & \texttt{claude-3-7-sonnet-20250219} & default: extended thinking disabled \\
Claude 4 Sonnet & \texttt{claude-sonnet-4-20250514} & default: extended thinking disabled \\
Gemini 2.5 Flash & \texttt{gemini-2.5-flash} & default: dynamic (adaptive) thinking \\
Gemini 2.5 Pro & \texttt{gemini-2.5-pro} & default: dynamic (adaptive) thinking \\
GPT-4o & \texttt{gpt-4o} & reasoning effort not applicable \\
GPT-5 & \texttt{gpt-5} & default: medium \\
GPT-o3 & \texttt{o3} & default: medium \\
\midrule
Claude Haiku 4.5 & \texttt{claude-haiku-4-5} & \texttt{thinking.type=enabled; budget\_tokens=8192} \\
Claude Sonnet 5 & \texttt{claude-sonnet-5} & \texttt{thinking.type=adaptive} \\
Claude Opus 4.8 & \texttt{claude-opus-4-8} & \texttt{thinking.type=adaptive} \\
GPT-5.6 Sol & \texttt{gpt-5.6-sol} & \texttt{reasoning\_effort=medium} \\
GPT-5.6 Terra & \texttt{gpt-5.6-terra} & \texttt{reasoning\_effort=medium} \\
GPT-5.6 Luna & \texttt{gpt-5.6-luna} & \texttt{reasoning\_effort=medium} \\
Gemini 3.5 Flash & \texttt{gemini-3.5-flash} & \shortstack[l]{\texttt{generationConfig.}\\\texttt{thinkingConfig.thinkingLevel=medium}} \\
Gemini 3.1 Pro & \texttt{gemini-3.1-pro-preview} & \shortstack[l]{\texttt{generationConfig.}\\\texttt{thinkingConfig.thinkingLevel=medium}} \\
\bottomrule
\end{tabular}
\end{table*}

\subsection{Hyperparameters}
\paragraph{VLM API Calls (Single-Agent \& Multi-Agent)}
The request-side sampling parameters used for the reported model columns are summarized below. ``Omitted'' means that the field was not sent and the provider default applied. The output limit includes any hidden thinking tokens and is distinct from the reasoning controls in the preceding section. For Claude it is sent as \texttt{max\_tokens}; for Gemini it is sent as \texttt{generationConfig.maxOutputTokens}. Single-agent and multi-agent temperatures are denoted by S and M, respectively.

{\footnotesize
\setlength{\tabcolsep}{3pt}
\renewcommand{\arraystretch}{1.08}
\begin{center}
\begin{tabular}{@{}p{0.4\columnwidth}p{0.20\columnwidth}p{0.30\columnwidth}@{}}
\toprule
\textbf{Models} & \textbf{Output cap} & \textbf{Temp.} \\
\midrule
Claude Sonnet 3.7 / 4 & \texttt{4096} & \texttt{1.0} / omitted \\
GPT-4o / o3 / GPT-5 & omitted & omitted / omitted \\
Gemini 2.5 Flash / Pro & omitted & omitted / omitted \\
Claude 4.5 / 5 / 4.8 & \texttt{16000} & \texttt{1.0} / omitted \\
GPT-5.6 Sol / Terra / Luna & omitted & omitted / omitted \\
Gemini 3.5 Flash / 3.1 Pro & \texttt{65536} & \texttt{1.0} / omitted \\
\bottomrule
\end{tabular}
\end{center}
}
The larger Claude and Gemini limits provide room for both model-internal thinking and the final answer; they are ceilings rather than fixed token allocations, and only generated tokens are counted as usage.

\paragraph{ViewNavigator Framework}
We use two explicitly separated ViewNavigator configurations:

{\footnotesize
\setlength{\tabcolsep}{2pt}
\renewcommand{\arraystretch}{1.08}
\begin{center}
\begin{tabular}{@{}p{0.32\columnwidth}p{0.17\columnwidth}p{0.25\columnwidth}p{0.14\columnwidth}@{}}
\toprule
\textbf{Mode} & \textbf{Steps} & \textbf{Belief} & \textbf{Jitter} \\
\midrule
Budget-matched & \texttt{6} & uniform & \texttt{0} \\
Full (unconstrained) & \texttt{10} & Wilson score & \texttt{5} \\
\bottomrule
\end{tabular}
\end{center}
}
Both configurations use the confidence threshold \texttt{tau = 0.6}. In the full configuration, \texttt{r\_az = r\_el = 5} degrees and \texttt{jitter\_size = 5} denotes five perturbed images in addition to the unperturbed view. The budget-matched configuration disables micro-jitter, yielding one image per step and at most six images in total.

\subsection{Prompts}
The exact prompts used in our experiments are provided below.  
Placeholders such as \texttt{\{central\_obj\_type\}} are filled dynamically during data generation.  

\begin{tcolorbox}[colback=MorandiBlue!10,colframe=MorandiBlue, title=Prompt for MultiView-Bench-3D and MultiView-Bench-2D Tasks (Single-agent):]
\begin{quote}
\small\ttfamily\raggedright
Look at this 3D scene carefully from different viewpoints. You can see several geometric objects and coordinate axes.

\medskip
\textbf{COORDINATE SYSTEM:}\\
- X-axis: RED rod, pointing to positive X direction\\
- Y-axis: GREEN rod, pointing to positive Y direction\\
- Z-axis: BLUE rod, pointing to positive Z direction\\
- Origin (0,0,0): YELLOW sphere, located at the center of the \{central\_obj\_type\}

\medskip
\textbf{TASK:}\\
Determine the relative position of the \{sampled\_obj\_type\} compared to the \{central\_obj\_type\} in terms of their geometric centers.

\medskip
\textbf{INSTRUCTIONS:}\\
1. Look at where the \{sampled\_obj\_type\} is positioned relative to the \{central\_obj\_type\}\\
2. For each axis, determine if the \{sampled\_obj\_type\} is in the positive (+) or negative (-) direction using the coordinate system shown in the images.\\
3. If objects appear at approximately the same level on an axis, use (0)

\medskip
\textbf{ANSWER FORMAT:}\\
Respond with exactly this format: \texttt{<answer>(±X, ±Y, ±Z)</answer>}\\
Examples: \texttt{<answer>(+X, -Y, +Z)</answer>} or \texttt{<answer>(-X, 0Y, -Z)</answer>} or \texttt{<answer>(0X, +Y, 0Z)</answer>}

\medskip
What is the relative position of the \{sampled\_obj\_type\} to the \{central\_obj\_type\}?
\end{quote}
\end{tcolorbox}

\newpage

\paragraph{Prompt for MultiView-Bench-2D Tasks (Multi-agent):}~\\

\begin{tcolorbox}[colback=MorandiBlue!10,colframe=MorandiBlue, title=Prompt for Front View (XZ plane)]
\begin{quote}
\small\ttfamily\raggedright
Look at this Front View (XZ plane) carefully. You can see several geometric objects and coordinate axes.

\medskip
\textbf{VIEW DESCRIPTION:}\\
This is the Front View (XZ plane), looking along the Y-axis.

\medskip
\textbf{COORDINATE SYSTEM:}\\
- X-axis: RED rod, pointing to positive X direction\\
- Z-axis: BLUE rod, pointing to positive Z direction\\
- Origin (0,0,0): YELLOW sphere, located at the center of the \{central\_obj\_type\}

\medskip
\textbf{TASK:}\\
Determine the relative position of the \{sampled\_obj\_type\} compared to the \{central\_obj\_type\} in terms of their geometric centers, focusing only on the X and Z axes visible in this view.

\medskip
\textbf{INSTRUCTIONS:}\\
1. Look at where the \{sampled\_obj\_type\} is positioned relative to the \{central\_obj\_type\}\\
2. For each visible axis (X, Z), determine if the \{sampled\_obj\_type\} is in the positive (+) or negative (-) direction using the coordinate system shown in the image.\\
3. If objects appear at approximately the same level on an axis, use (0)

\medskip
\textbf{ANSWER FORMAT:}\\
Respond with exactly this format for the X and Z axes: \texttt{<answer>(±X, ±Z)</answer>}\\
Examples: \texttt{<answer>(+X, -Z)</answer>} or \texttt{<answer>(0X, +Z)</answer>}

\medskip
What is the relative position of the \{sampled\_obj\_type\} to the \{central\_obj\_type\} in the X and Z axes?
\end{quote}
\end{tcolorbox}

\newpage

\begin{tcolorbox}[colback=MorandiBlue!10,colframe=MorandiBlue,title=Prompt for Side View (YZ plane)]
\begin{quote}
\small\ttfamily\raggedright
Look at this Side View (YZ plane) carefully. You can see several geometric objects and coordinate axes.

\medskip
\textbf{VIEW DESCRIPTION:}\\
This is the Side View (YZ plane), looking along the X-axis.

\medskip
\textbf{COORDINATE SYSTEM:}\\
- Y-axis: GREEN rod, pointing to positive Y direction\\
- Z-axis: BLUE rod, pointing to positive Z direction\\
- Origin (0,0,0): YELLOW sphere, located at the center of the \{central\_obj\_type\}

\medskip
\textbf{TASK:}\\
Determine the relative position of the \{sampled\_obj\_type\} compared to the \{central\_obj\_type\} in terms of their geometric centers, focusing only on the Y and Z axes visible in this view.

\medskip
\textbf{INSTRUCTIONS:}\\
1. Look at where the \{sampled\_obj\_type\} is positioned relative to the \{central\_obj\_type\}\\
2. For each visible axis (Y, Z), determine if the \{sampled\_obj\_type\} is in the positive (+) or negative (-) direction using the coordinate system shown in the image.\\
3. If objects appear at approximately the same level on an axis, use (0)

\medskip
\textbf{ANSWER FORMAT:}\\
Respond with exactly this format for the Y and Z axes: \texttt{<answer>(±Y, ±Z)</answer>}\\
Examples: \texttt{<answer>(+Y, -Z)</answer>} or \texttt{<answer>(0Y, +Z)</answer>}

\medskip
What is the relative position of the \{sampled\_obj\_type\} to the \{central\_obj\_type\} in the Y and Z axes?
\end{quote}
\end{tcolorbox}

\noindent 


\begin{tcolorbox}[colback=MorandiBlue!10,colframe=MorandiBlue,title=Prompt for Top View (XY plane)]
\begin{quote}
\small\ttfamily\raggedright
Look at this Top View (XY plane) carefully. You can see several geometric objects and coordinate axes.

\medskip
\textbf{VIEW DESCRIPTION:}\\
This is the Top View (XY plane), looking along the Z-axis from above.

\medskip
\textbf{COORDINATE SYSTEM:}\\
- X-axis: RED rod, pointing to positive X direction\\
- Y-axis: GREEN rod, pointing to positive Y direction\\
- Origin (0,0,0): YELLOW sphere, located at the center of the \{central\_obj\_type\}

\medskip
\textbf{TASK:}\\
Determine the relative position of the \{sampled\_obj\_type\} compared to the \{central\_obj\_type\} in terms of their geometric centers, focusing only on the X and Y axes visible in this view.

\medskip
\textbf{INSTRUCTIONS:}\\
1. Look at where the \{sampled\_obj\_type\} is positioned relative to the \{central\_obj\_type\}\\
2. For each visible axis (X, Y), determine if the \{sampled\_obj\_type\} is in the positive (+) or negative (-) direction using the coordinate system shown in the image.\\
3. If objects appear at approximately the same level on an axis, use (0)

\medskip
\textbf{ANSWER FORMAT:}\\
Respond with exactly this format for the X and Y axes: \texttt{<answer>(±X, ±Y)</answer>}\\
Examples: \texttt{<answer>(+X, -Y)</answer>} or \texttt{<answer>(0X, +Y)</answer>}

\medskip
What is the relative position of the \{sampled\_obj\_type\} to the \{central\_obj\_type\} in the X and Y axes?
\end{quote}
\end{tcolorbox}

\newpage 
\paragraph{Prompts for ViewNavigator:}~\\

\subparagraph{VLM Perception Module Prompts}~\\
\noindent 
\begin{tcolorbox}[colback=MorandiBlue!10,colframe=MorandiBlue,title=\texttt{<VLM\_SYSTEM\_PROMPT>}]
\begin{quote}
\small\ttfamily\raggedright
You are a precise vision judge. The image shows colored world axes:

\medskip
\textbf{COORDINATE SYSTEM:}\\
- X-axis: RED rod, pointing to positive X direction\\
- Y-axis: GREEN rod, pointing to positive Y direction\\
- Z-axis: BLUE rod, pointing to positive Z direction\\
- Origin (0,0,0): YELLOW sphere, located at the center of the \{central\_object\}\\
- Both CENTRAL and TARGET objects have the same scale in X, Y, Z dimensions

\medskip
\textbf{TASK:}\\
Determine the relative position of the \{target\_object\} compared to the \{central\_object\} in terms of their geometric centers.

\medskip
\textbf{INSTRUCTIONS:}\\
1. Only focus on axis \{axis\} for this view and only give answer for these axes.\\
2. Compare the TARGET center to the CENTRAL center along each of \{axis\}:\\
\quad - "+" if TARGET lies in the positive direction\\
\quad - "-" if in the negative direction\\
\quad - "0" if approximately equal (centers aligned along that axis)\\
3. Wrap your full step-by-step reasoning in \texttt{<think>...</think>}.\\
4. Then emit exactly one line, wrapped in \texttt{<answer>...</answer>}, listing only axes \{axis\} with their sign or 0.

\medskip
Do \textbf{not} include any extra text or prose
\end{quote}
\end{tcolorbox}

\noindent 
\begin{tcolorbox}[colback=MorandiBlue!10,colframe=MorandiBlue,title=\texttt{<VLM\_USER\_PROMPT>}]
\begin{quote}
\small\ttfamily\raggedright
CENTRAL: \{central\}\\
TARGET: \{target\}

\medskip
Return exactly:\\
\texttt{<think>...step-by-step reasoning...</think>}\\
\texttt{<answer>(±X, ±Y, ±Z)</answer>}

\medskip
\textbf{Valid examples:}\\
\quad \texttt{<think>I see red and green axes...</think>}\\
\quad \texttt{<answer>(+X, -Y)</answer>}

\medskip
\quad \texttt{<think>Blue axis only is clear...</think>}\\
\quad \texttt{<answer>(-Z)</answer>}
\end{quote}
\end{tcolorbox}

\newpage

\subparagraph{LLM Planner Module Prompts}~\\

\noindent 
\begin{tcolorbox}[breakable,colback=MorandiBlue!10,colframe=MorandiBlue,title=\texttt{<LLM\_SYSTEM\_PROMPT>}]
\begin{quote}
\small\ttfamily\raggedright
You control a camera in a 3D scene. Your goal is to decide the signs (+,0,-) of TARGET relative to CENTRAL on axes X,Y,Z by choosing successive viewpoints.

\medskip
\textbf{Camera Coordinate System:}\\
- Azimuth 0\textdegree{}: X-axis points towards viewer, Y-axis points right\\
- As azimuth increases (clockwise rotation):\\
\quad - Azimuth 90\textdegree{}: Y-axis points towards viewer, X-axis points left\\
\quad - Azimuth 180\textdegree{}: X-axis points away from viewer, Y-axis points left\\
\quad - Azimuth 270\textdegree{}: Y-axis points away from viewer, X-axis points right\\
- Elevation 0\textdegree{}: Camera views from directly above (top-down)\\
- Elevation 90\textdegree{}: Camera views from horizontal level\\
- Elevation 180\textdegree{}: Camera views from directly below (bottom-up)

\medskip
On every turn you will receive:\\
- threshold $\tau$ (a float in [0,1])\\
- belief\_state:\\
\quad \{\\
\quad\quad "X": \{"+" : p\_plus, "0": p\_zero, "-" : p\_minus\},\\
\quad\quad "Y": \{...\},\\
\quad\quad "Z": \{...\}\\
\quad \}\\
- history: a list of previously checked views, each entry:\\
\quad \{\\
\quad\quad "view": \{"az": az\_deg, "el": el\_deg\},\\
\quad\quad "answer": "(±X, ±Y, ±Z)" or shorter,\\
\quad\quad "confidence": \{"X":cX, "Y":cY, "Z":cZ\}\\
\quad \}

\medskip
If \textbf{all} axes have max(belief) $\geq \tau$, you should stop. Otherwise choose the next best view. Note that you can revisit some views to strengthen your belief. You should also decide which axes you want to focus on in a view. For example, if you choose a view that shows the XY plane, then you should focus on only the X axis and Y axis or even just focus on X or Y axis.

\medskip
Note that the confidence score represents the reliability of the answer got from that view. Zero confidence score for a view indicates that the relative position is not clear revealed through that view.

\medskip
\textbf{Rules:}\\
1. Wrap your internal reasoning in \texttt{<think>...</think>}\\
2. Then emit exactly one \texttt{<answer>...</answer>} containing \textbf{only} this JSON:\\
\{\\
\quad "action": "CAPTURE"|"STOP",\\
\quad "view": \{"az": <number>, "el": <number>\} | null,\\
\quad "axis": ["X", "Y"]\\
\}

\medskip
No extra text or fields.
\end{quote}
\end{tcolorbox}

\noindent 
\begin{tcolorbox}[colback=MorandiBlue!10,colframe=MorandiBlue,title=\texttt{<LLM\_FIRST\_TURN\_PROMPT>}]
\begin{quote}
\small\ttfamily\raggedright
\# First turn (no belief\_state or history)

\medskip
Task: find (±X,±Y,±Z) for TARGET=\{target\} vs CENTRAL=\{central\}.\\
Threshold $\tau$ = \{tau\}.

\medskip
Propose your initial viewpoint.

\medskip
Respond with:\\
\texttt{<think>...</think>}\\
\texttt{<answer>\{\{}\\
\quad "action": "CAPTURE",\\
\quad "view": \{\{"az": <num>, "el": <num>\}\},\\
\quad "axis": ["axes to focus on for this view"]\\
\}\}</answer>
\end{quote}
\end{tcolorbox}

\begin{tcolorbox}[colback=MorandiBlue!10,colframe=MorandiBlue,title=\texttt{<LLM\_INTERMEDIATE\_TURN\_PROMPT>}]
\begin{quote}
\small\ttfamily\raggedright
\# Subsequent turn

\medskip
Threshold $\tau$ = \{tau\}\\
belief\_state = \{belief\_state\}\\
history = \{history\}

\medskip
Decide whether to STOP or pick another view.

\medskip
Respond with:\\
\texttt{<think>...</think>}\\
\texttt{<answer>\{\{}\\
\quad "action": "CAPTURE"|"STOP",\\
\quad "view": \{\{"az": <num>, "el": <num>\}\} | null,\\
\quad "axis": ["axes to focus on for this view"] | null\\
\}\}</answer>
\end{quote}
\end{tcolorbox}

\newpage 

\section{Examples of More Spatial Tasks \label{more_spatial_tasks}}

\begin{figure}[h]
\begin{center}
\includegraphics[width=\textwidth]{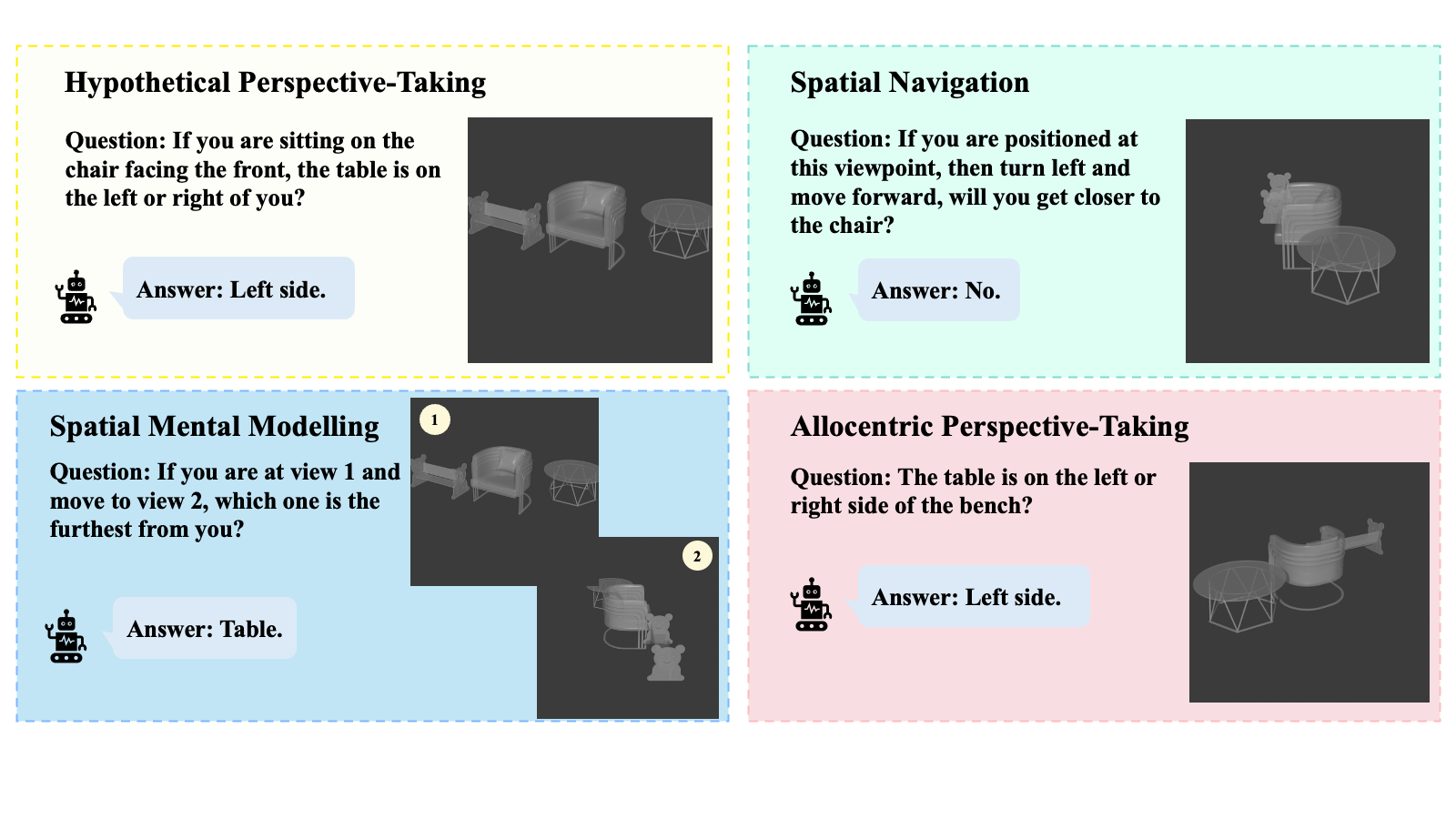}
\end{center}
\caption{Examples of more spatial tasks that can be created using our data generation pipeline.}
\label{fig:more_possible_tasks}
\end{figure}

\newpage 

\section{Failure Cases}
In this section, we present a glimpse of failure cases we observed during the evaluations.

\subsection{3D DoF=3 (Claude 3.7 Sonnet)}

In this challenging 3D spatial reasoning task, the model correctly identifies the relationships on the X and Y axes but fails to accurately determine the relative position on the Z-axis.
\begin{itemize}
    \item \textbf{Ground Truth:} \texttt{(-X, +Y, +Z)}
    \item \textbf{Model's Answer:} \texttt{(-X, +Y, 0Z)}
\end{itemize}

\begin{figure}[h!]
    \centering
    \begin{subfigure}[b]{0.25\textwidth}
        \centering
        \includegraphics[width=\textwidth]{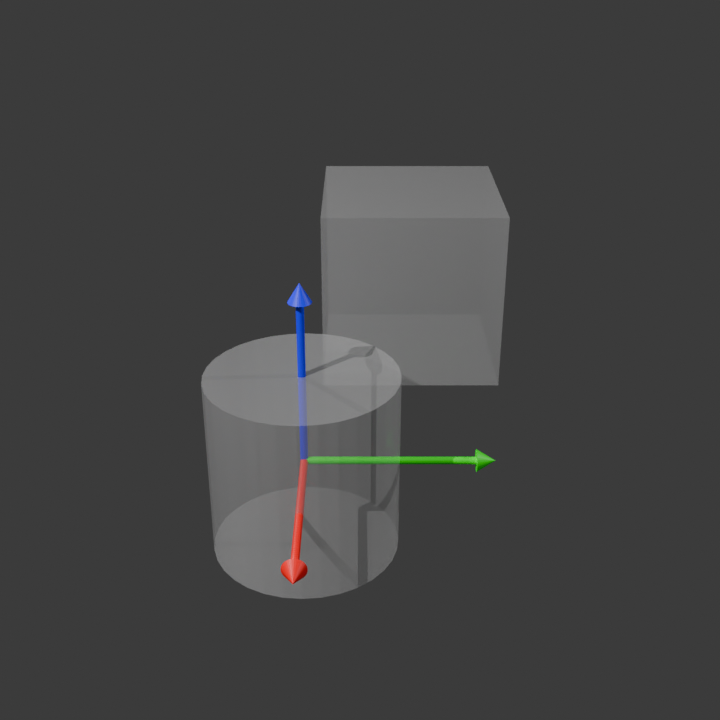}
    \end{subfigure}
    \hfill
    \begin{subfigure}[b]{0.25\textwidth}
        \centering
        \includegraphics[width=\textwidth]{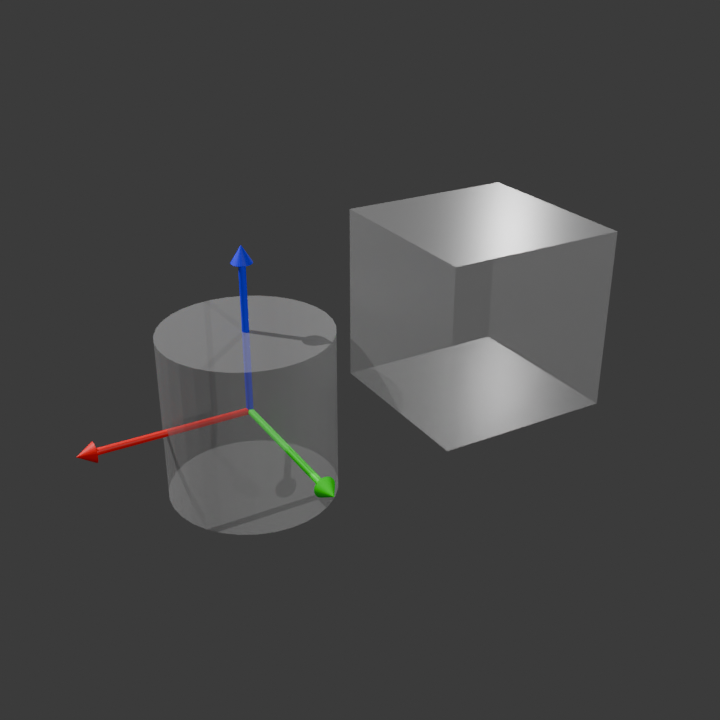}
    \end{subfigure}
    \hfill
    \begin{subfigure}[b]{0.25\textwidth}
        \centering
        \includegraphics[width=\textwidth]{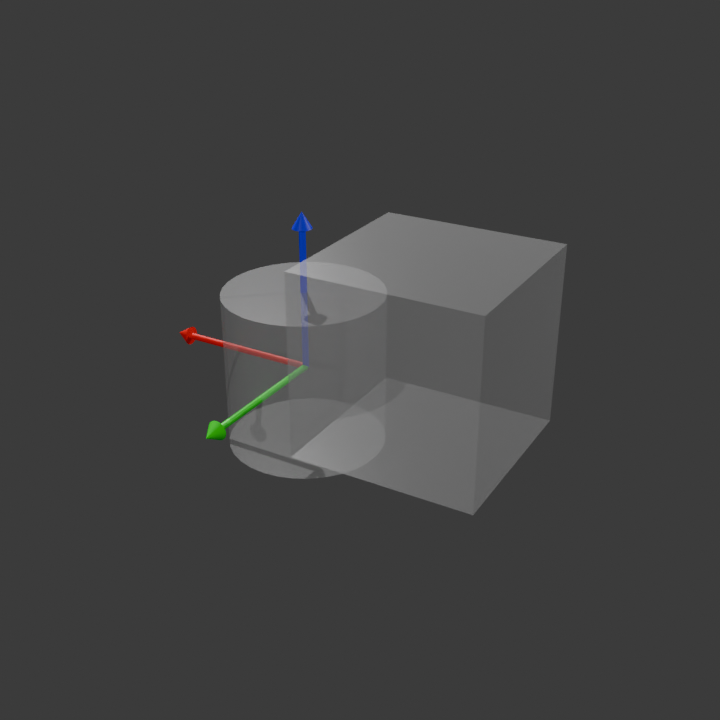}
    \end{subfigure}
    
    \par\medskip 
    
    \begin{subfigure}[b]{0.25\textwidth}
        \centering
        \includegraphics[width=\textwidth]{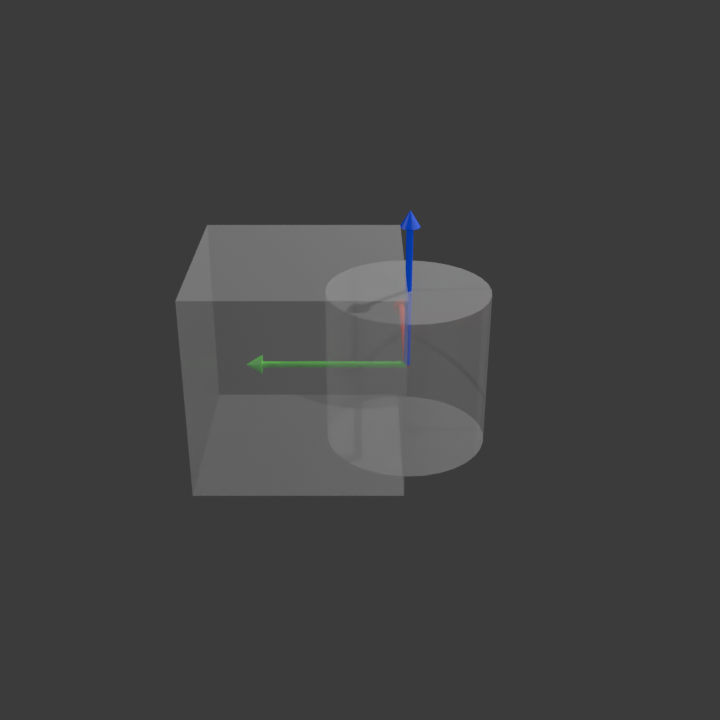}
    \end{subfigure}
    \hfill
    \begin{subfigure}[b]{0.25\textwidth}
        \centering
        \includegraphics[width=\textwidth]{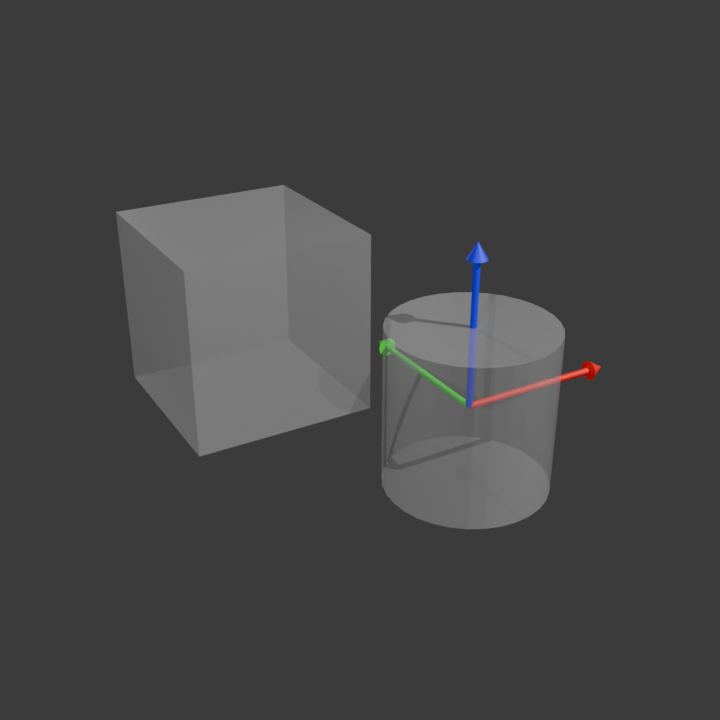}
    \end{subfigure}
    \hfill
    \begin{subfigure}[b]{0.25\textwidth}
        \centering
        \includegraphics[width=\textwidth]{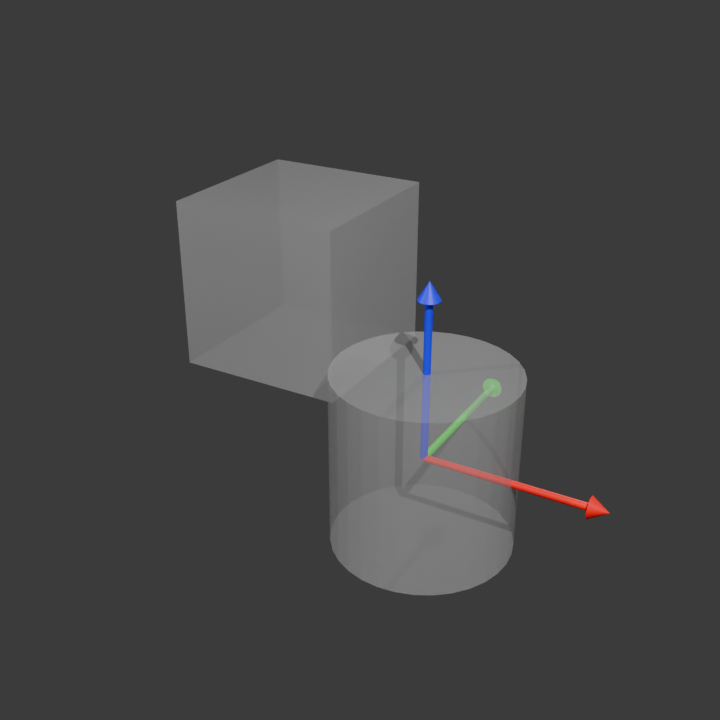}
    \end{subfigure}
    \caption{Six isometric views provided to the model for the 3D DoF=3 task.}
    \label{fig:3d_dof3_case}
\end{figure}

\textit{\textbf{Reasoning and Answer:}}
\noindent
\begin{tcolorbox}[colback=MorandiBlue!10,colframe=MorandiBlue]
Looking at the different viewpoints of the 3D scene, I can analyze the relative position of the cube compared to the cylinder.

From the images, I can see that:
\begin{itemize}
    \item X-axis: The cube appears to be positioned mainly in the negative X direction relative to the cylinder (to the left in several views)
    \item Y-axis: The cube appears to be positioned in the positive Y direction relative to the cylinder (forward in several views)
    \item Z-axis: The cube and cylinder appear to be at approximately the \textbf{\textcolor{red}{same height} on the Z-axis}
\end{itemize}
\texttt{<answer>}(-X, +Y, \textbf{\textcolor{red}{0Z}})\texttt{</answer>}
\end{tcolorbox}

\newpage
\subsection{2D Colors (Claude 4 Sonnet)}

In this task, the model is provided with three canonical 2D views (front, side, and top). The model demonstrates a classic failure in view integration. It correctly processes the x-axis relationship (\texttt{-X}), which is salient in the top and front views. However, when trying to determine depth (Y-axis from the side/top views) and height (Z-axis from the front/side views), it incorrectly concludes the objects are at the "same level."

\begin{itemize}
    \item \textbf{Ground Truth:} \texttt{(-X, +Y, +Z)}
    \item \textbf{Model's Answer:} \texttt{(-X, 0Y, 0Z)}
\end{itemize}

\begin{figure}[h]
    \centering
    \begin{subfigure}[b]{0.3\textwidth}
        \centering
        \includegraphics[width=\textwidth]{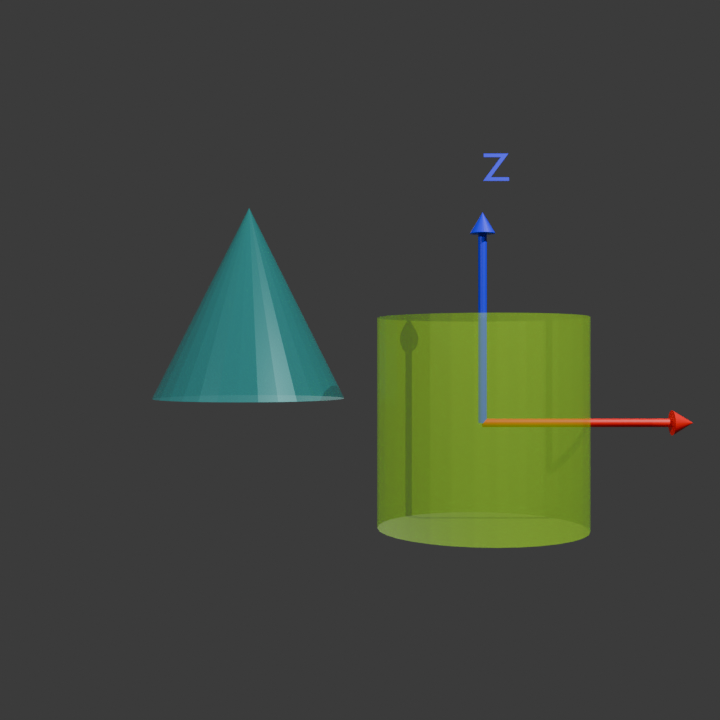}
        \caption{Front view}
    \end{subfigure}
    \hfill
    \begin{subfigure}[b]{0.3\textwidth}
        \centering
        \includegraphics[width=\textwidth]{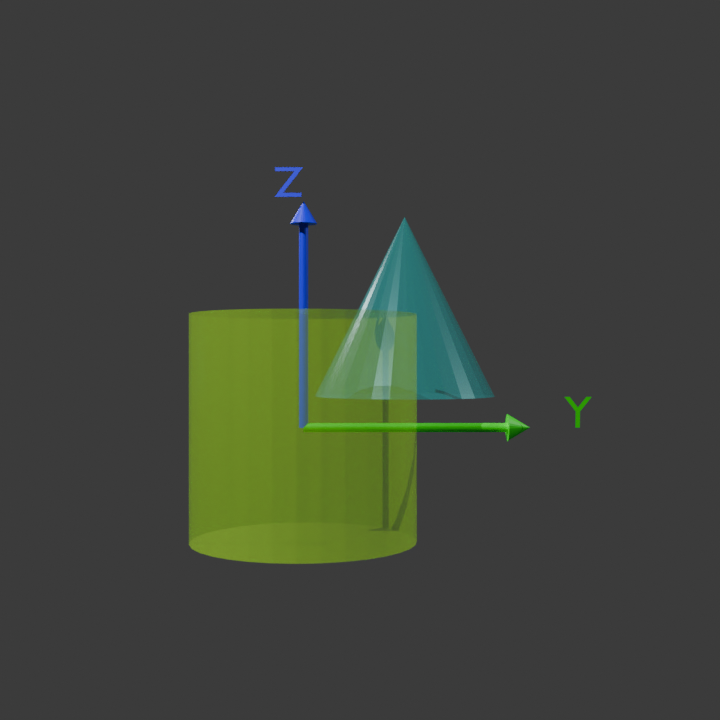}
        \caption{Side view}
    \end{subfigure}
    \hfill
    \begin{subfigure}[b]{0.3\textwidth}
        \centering
        \includegraphics[width=\textwidth]{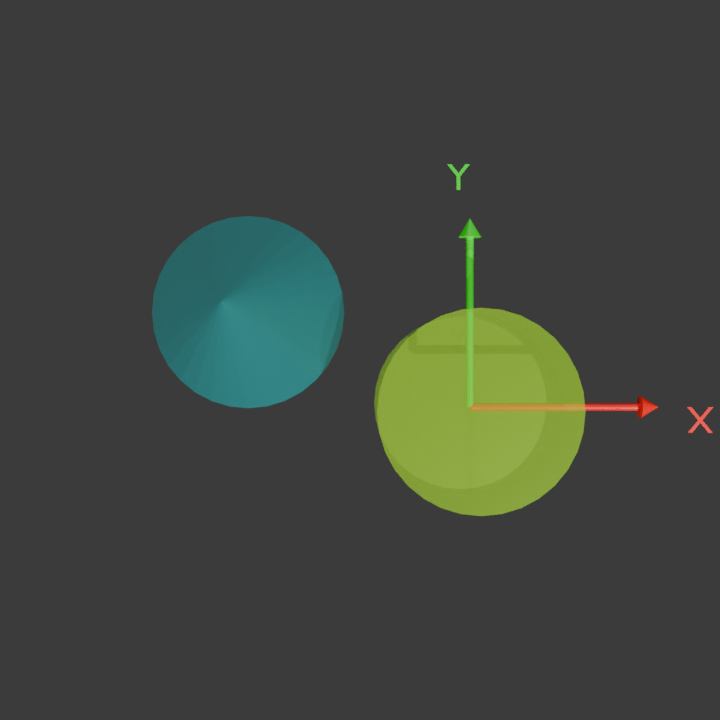}
        \caption{Top view}
    \end{subfigure}
    \caption{Example task of 2D three view with colors.}
\end{figure}

\textit{\textbf{Reasoning and Answer:}}
\noindent
\begin{tcolorbox}[colback=MorandiBlue!10,colframe=MorandiBlue]

Based on examining the 3D scene from multiple viewpoints, I can determine the relative position of the teal cone compared to the lime cylinder.

Looking at all three viewpoints:
\begin{itemize}
    \item X-axis: The cone appears to be to the left of the cylinder (negative X direction)
    \item Y-axis: The cone appears to be at approximately the \textbf{\textcolor{red}{same level}} as the cylinder on the Y-axis
    \item Z-axis: The cone appears to be at approximately the \textbf{\textcolor{red}{same height}} as the cylinder on the Z-axis 
\end{itemize}
\texttt{<answer>}(-X, \textbf{\textcolor{red}{0Y, 0Z}})\texttt{</answer>}

\end{tcolorbox}

\newpage

\subsection{ViewNavigator (GPT-5): VLM Perception Failure}

This case illustrates a failure where the agent incorrectly determines the position on the Y-axis, leading to an incorrect final answer. The failure stems from a critical misperception by the VLM, which then misleads the LLM planner. This single error propagates through the agent's belief state, causing the LLM planner to develop confirmation bias and eventually accept the wrong conclusion.
\begin{itemize}
    \item \textbf{Ground Truth:} \texttt{(-X, -Y, +Z)}
    \item \textbf{Agent's Final Answer:} \texttt{(-X, 0Y, +Z)}
\end{itemize}

\begin{figure}[h]
    \centering
    \includegraphics[width=0.3\linewidth]{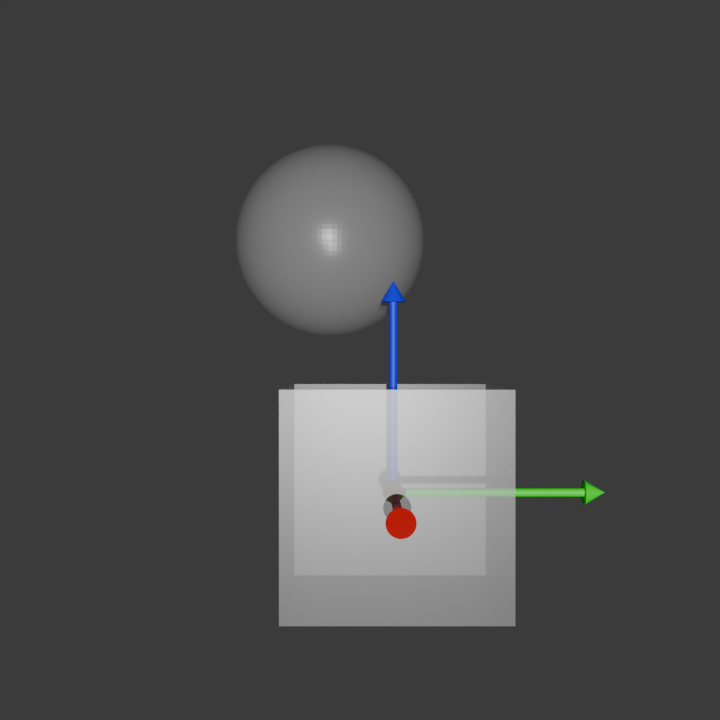}
    \caption{Image observed by the VLM from a side view (\texttt{az=0, el=90}).}
    \label{fig:vlm_fail}
\end{figure}

\textit{\textbf{Reasoning and Answer:}}
\noindent
\begin{tcolorbox}[colback=MorandiBlue!10,colframe=MorandiBlue]
\texttt{<think>}Identify axes: green is +Y to the right; blue is +Z upward. Compare centers: the sphere is directly above the cube with \textbf{\textcolor{red}{no horizontal (Y) offset}}, so along Y the centers align (0). Along Z, the sphere lies above the cube center, so it is in the positive Z direction (+).\texttt{</think>}
\texttt{<answer>}(\textbf{\textcolor{red}{0Y}}, +Z)\texttt{</answer>}
\end{tcolorbox}

\newpage
\subsection{ViewNavigator (Gemini 2.5 Flash): LLM Planning Failure}

This case demonstrates a fundamental failure in the LLM's spatial reasoning, where there is a disconnect between its stated goal and its proposed action. It correctly identifies the need for a view of the XZ plane but incorrectly believes that an elevation of 90\textdegree{} will achieve this. A view with an elevation of 90\textdegree{} is a top-down view, which clearly displays the XY plane while making it impossible to accurately judge relative positions along the X-axis. The model's stated goal is contradicted by its proposed action, highlighting a limited capacity to translate a desired outcome into the correct geometric parameters.

\begin{figure}[h]
    \centering
    \includegraphics[width=0.3\linewidth]{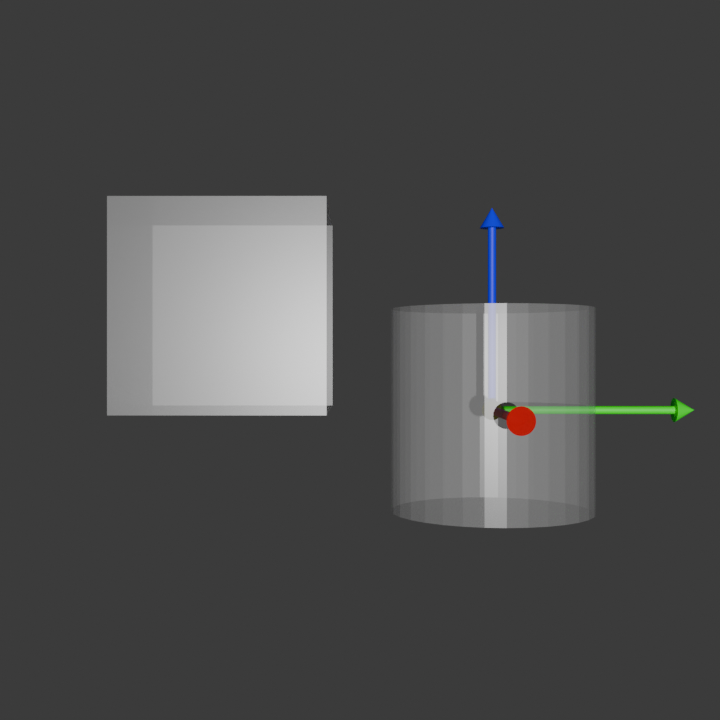}
    \caption{The view generated by the LLM's proposal (\texttt{az=0, el=90}).}
    \label{fig:llm_fail}
\end{figure}

\textit{\textbf{Reasoning and Planning:}}
\noindent
\begin{tcolorbox}[colback=MorandiBlue!10,colframe=MorandiBlue]
\texttt{<think>}
... A view with azimuth 0\textdegree{} and elevation 90\textdegree{} would provide a side view focusing on the XZ plane, which might help clarify the X axis. This view also gives us a chance to improve our understanding of the Z axis.
Therefore, I will choose a view with azimuth 0\textdegree{} and elevation 90\textdegree{}, focusing on the X and Z axes.
\texttt{</think>}
\texttt{<answer>}
{
  "action": "CAPTURE",
  "view": {"az": 0, "el": 90},
  "axis": ["X", "Z"]
}
\texttt{</answer>}
\end{tcolorbox}

\newpage
\begin{figure*}[!t]
    \subsection{ViewNavigator with more budget}
    \label{sec:full_agent}
    The main text reports ViewNavigator under a budget-matched constraint, so that its gains cannot be attributed to extra test-time computation. Here we let the LLM planner actively select up to 10 viewpoints and the VLM queries multiple micro-jittered observations per viewpoint. This larger test-time budget further amplifies the improvements over the budget-matched setting for GPT-5 but remains comparable to or worse than the budget-matched setting for other weaker base models.

    Notably, for GPT-5, the micro-jittering strategy proves crucial: by sampling slightly perturbed viewpoints, we expose inconsistencies in VLM predictions that would otherwise go undetected. This suggests that VLM spatial reasoning is highly sensitive to subtle viewpoint changes---a vulnerability that ViewNavigator exploits to improve robustness.
    \par\medskip
    \centering
    \includegraphics[width=\textwidth]{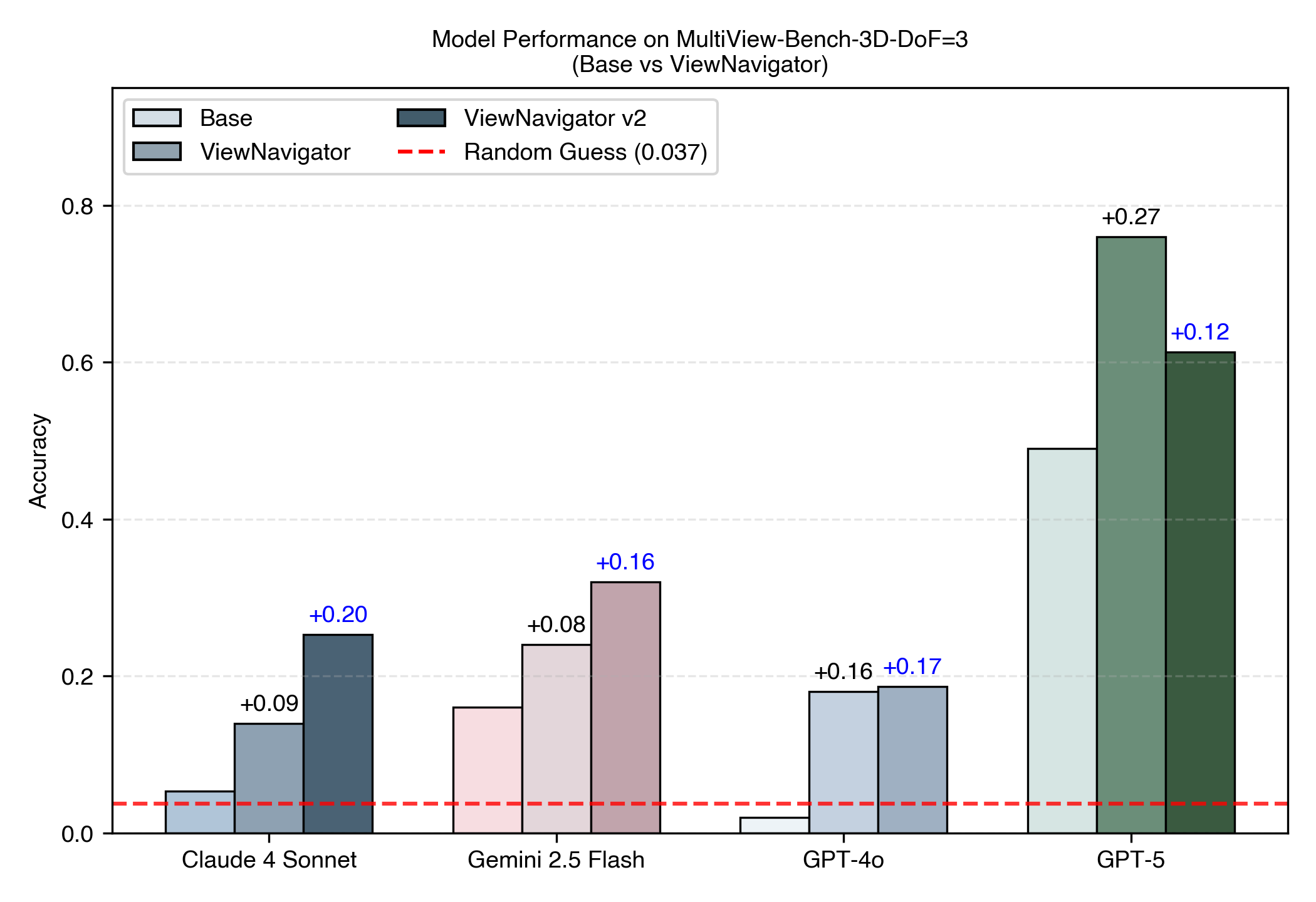}
    \caption{ViewNavigator comparison on the 3D DoF$=3$ task for four base models. Bars show the base model, the full unconstrained agent, and the budget-matched six-view configuration. The full agent may actively select up to 10 viewpoints and query micro-jittered observations at each step, whereas the six-view configuration matches the budget used in the main paper.}
    \label{fig:viewnavigator_full}
\end{figure*}
    
}{}
\makeatletter
\ifthenelse{\equal{\compileversion}{appendixonly}}{%
    \vbox{%
        \hsize\textwidth
        \linewidth\hsize
        \vskip 0.1in
        \centering
        {\LARGE\bfseries [Appendix] \@title\par} 
        \vskip 0.1in
      }
    
    
    \bibliography{references}
}{}
\makeatother


\end{document}